\documentclass{article}

\usepackage{iclr2027_conference,times}

\usepackage{amsmath,amsfonts,bm}

\def\eqref#1{equation~\ref{#1}}

\def\1{\bm{1}}

\DeclareMathAlphabet{\mathsfit}{\encodingdefault}{\sfdefault}{m}{sl}
\SetMathAlphabet{\mathsfit}{bold}{\encodingdefault}{\sfdefault}{bx}{n}

\usepackage{amsmath}
\usepackage{amssymb}
\usepackage{booktabs}
\usepackage{xcolor}
\usepackage{multirow}
\usepackage{microtype}
\usepackage{graphicx}
\usepackage{subcaption}
\usepackage{url}
\usepackage{wrapfig}
\usepackage{tabularx}
\usepackage{algorithm}
\usepackage{algpseudocode}
\usepackage{float}
\usepackage[hidelinks]{hyperref}

\newif\ifswamrevisionmarked
\swamrevisionmarkedtrue

\newcommand{\swamrev}[1]{\ifswamrevisionmarked\textcolor{blue}{#1}\else#1\fi
}

\newenvironment{swamrevision}{\ifswamrevisionmarked\color{blue}\fi
}{}
\swamrevisionmarkedfalse

\graphicspath{{figures/}}

\newcommand{\method}{Sparse-WAM}

\title{Sparse-WAM: Accelerating World Action Models
via Action-Guided Sparse Imagination}

\author{\normalfont
  \textbf{Xinling Xie}\textsuperscript{1,2,*,\textdagger}\quad
  \textbf{Haodong Wang}\textsuperscript{2,*}\quad
  \textbf{Mi Jiazhi}\textsuperscript{2}\quad
  \textbf{Zhiming Liu}\textsuperscript{3}\quad
  \textbf{Zicong Hong}\textsuperscript{4}\\
  \textbf{Xiaoyi Pang}\textsuperscript{2}\quad
  \textbf{Qianli Liu}\textsuperscript{2}\quad
  \textbf{Yangjia Hu}\textsuperscript{2}\quad
  \textbf{Zhengyang Yan}\textsuperscript{2}\quad
  \textbf{Ying Chen}\textsuperscript{2}\quad
  \textbf{Song Guo}\textsuperscript{2}\\[4pt]
  \textsuperscript{1}NJU\quad
  \textsuperscript{2}HKUST\quad
  \textsuperscript{3}HIT\quad
  \textsuperscript{4}EPFL\\[2pt]
  {\small\textsuperscript{*}Equal contribution.\quad
   \textsuperscript{\textdagger}Work done during an internship at HKUST.}
}

\iclrfinalcopy

\begin{document}

\maketitle
\lhead{}
\renewcommand{\headrulewidth}{0.4pt}

\vspace{-10pt}
\begin{abstract}
World-action models (WAMs) leverage pretrained video models to improve generalization in robot control by jointly predicting future visual states and actions. This capability comes at a substantial inference cost, as dense future-frame tokens are repeatedly processed during denoising. Prior methods address this by token pruning that prioritizes visual fidelity to reduce denoising costs in video diffusion models. However, these methods do not use action relevance to determine which future-frame tokens to retain during joint denoising in WAMs. In this paper, we propose \textbf{Sparse-WAM}, a training-free framework for \emph{action-guided sparse imagination} that selectively processes future-frame tokens to accelerate WAM inference. We observe substantial overlap in the spatial distribution of attention from action tokens to future-frame tokens (action-to-future attention) between consecutive denoising steps, despite continued updates to the future representations. Motivated by this, we develop \emph{Action-Guided Token Selection} to retain frame-specific action-relevant regions together with cross-frame context. 
However, a naive implementation can incur attention-scoring
and token-packing overhead that offsets the computational
savings from pruning.
We therefore introduce \textbf{Pilot}, an efficient engine
that reduces sparse inference overhead through lightweight
scoring and cross-step reuse of token selections.
On LIBERO with FastWAM-Joint and RoboLab-120 with Cosmos 3 Edge,
\textbf{Sparse-WAM} achieves inference speedups of approximately
$2.0\times$ and $1.8\times$, respectively, over dense eager
inference on an NVIDIA RTX 4090, while largely preserving
task performance.
\footnote{Project page: \url{https://rockylix.github.io/Sparse-WAM.github.io/}}
\end{abstract}

\section{Introduction}
\label{sec:introduction}

World action models (WAMs) have emerged as a promising paradigm for robotic control. Recent WAMs such as DreamZero~\citep{ye2026dreamzero} and Cosmos 3~\citep{agarwal2026cosmos3} jointly denoise future visual states and actions, leveraging spatiotemporal priors to improve generalization and robustness~\citep{zhang2026wamrobustness}. 
However, explicitly generating high-resolution future frames introduces a large number of visual tokens into each denoising step. Since the cost of each denoising step grows significantly
with the number of processed tokens~\citep{anagnostidis2025flexidit},
repeatedly processing these future-frame tokens increases
inference latency and limits control frequency~\citep{guo2024prediction}.

\begin{figure}[t]
  \centering
  \includegraphics[width=0.9\textwidth]{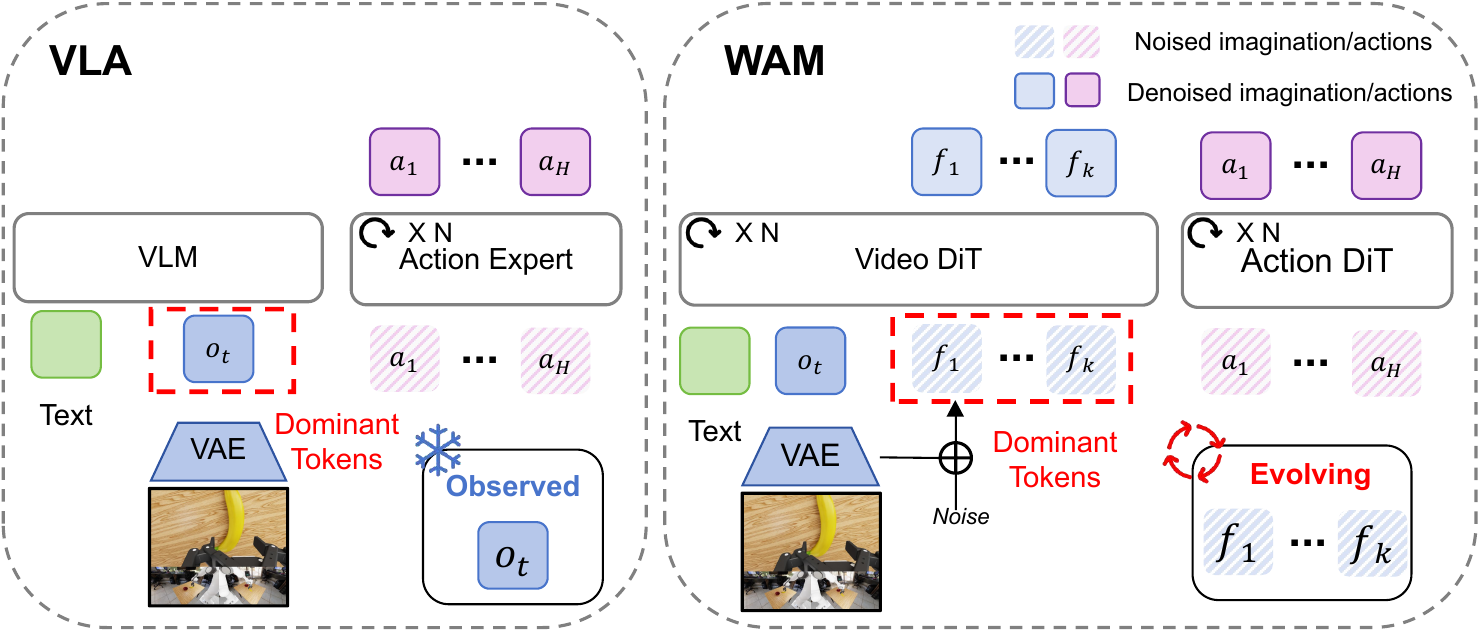}
  \caption{Representative architectures of VLA and WAM. In WAM, noisy imagination tokens constitute the majority of the visual-action sequence and    evolve throughout denoising.}
  \label{fig:comparison}
\end{figure}

To reduce this repeated computation, existing methods either
remove imagination at inference or compress future-frame tokens.
The former uses future imagination during training but omits
explicit future prediction during action
generation~\citep{yuan2026fastwam,li2026metis}.
The latter retains future prediction with fewer or lower-fidelity
visual tokens~\citep{chen2026lawam,li2026efficientwam},
but still devotes computation to regions with limited
relevance to actions.

Token pruning offers a fine-grained means of reducing the cost of
processing dense imagination tokens by selectively retaining
a subset for computation.
This paradigm has been explored in VLAs to retain useful information
from observed inputs~\citep{xu2025vlacache,wang2026specprune}
and in video generation to preserve visual
quality~\citep{zou2025accelerating,feng2026worldcache}.
In WAMs, however, future-frame tokens are initialized from noise
and jointly denoised with action tokens, making action-relevant
regions difficult to determine in advance.

To understand how action-relevant regions evolve during denoising, 
we examine attention from action tokens to future-frame tokens
(see \autoref{sec:observations}).
We observe that, despite continued changes in future representations,
the regions attended to by action tokens exhibit substantial
spatial overlap between consecutive denoising steps.
This consistency offers an opportunity to reuse token selections
across steps, but does not imply that the relevant regions remain
unchanged throughout denoising.
These observations motivate using action-to-future attention to select future regions for sparse computation and reusing the selections across denoising steps. However, attention scoring and token reorganization introduce additional overhead that can offset the computational savings. The core challenge is therefore to identify useful future regions while keeping the cost of selection and sparse execution low.

To address this challenge, we introduce \textbf{\method{}}, a training-free framework for action-guided sparse imagination in WAMs. It uses action-to-future attention to retain frame-specific regions together with shared spatial context. An efficient execution engine combines lightweight scoring with cross-step selection reuse to reduce the cost of joint visual–action inference.

We summarize our contributions as follows:
\begin{itemize}
    \item We reveal substantial spatial overlap in action-to-future
    attention between consecutive denoising steps, despite continued
    changes in future representations.
    This finding motivates reusing action-guided token selections
    during joint denoising.

    \item We propose \textbf{\method{}}, which uses temporally aligned
    action-to-future attention to select frame-specific regions
    together with shared spatial context.
    It concentrates Transformer computation on selected future
    tokens while retaining all observation and action tokens.

\item We develop \textbf{Pilot}, an efficient engine for online
token selection and sparse execution.
It combines lightweight attention profiling with cross-step
reuse of selected positions and packing metadata,
reducing the overhead of action-guided sparse inference.

    \item We evaluate \method{} on three WAMs across
    LIBERO~\citep{liu2023libero},
    RoboLab-120~\citep{yang2026robolab},
    LIBERO-Plus~\citep{fei25libero-plus},
    and real-world robotic tasks.
    On LIBERO, \method{} achieves a $1.98\times$ speedup
    over dense eager inference with a $0.30$ percentage-point
    decrease in average task success.
\end{itemize}

\section{Related Work}
\label{sec:related-work}

\paragraph{Efficient World Action Models.}
Repeated denoising of future-frame tokens makes diffusion-based
WAM inference computationally
expensive~\citep{ye2026dreamzero,li2025unified,bi2025motus}.
One line of work retains future prediction during training
but generates actions directly from learned world representations
at inference~\citep{yuan2026fastwam,ye2026gigaworld,li2026metis,li2026lightwam}.
Another preserves inference-time future modeling while reducing
its cost through compact latent subgoals, lower-resolution
representations, asymmetric denoising, or feature
reuse~\citep{chen2026lawam,li2026efficientwam,zhao2026fasterwam}.
Our method retains joint visual--action denoising and selectively
allocates future-frame token computation according to action relevance, without additional model training.

\paragraph{Token-Level Caching and Pruning.}
Token caching and pruning reduce inference cost through feature
reuse~\citep{lin2026kvdrive} and token removal.
VLA methods use observation similarity and task or action
cues~\citep{pei2026actionaware,liu2025vlapruner,ma2026safepruner},
as exemplified by VLA-Cache and
SpecPrune-VLA~\citep{xu2025vlacache,wang2026specprune}.
In video generation and diffusion world models~\citep{miao2026onlinewm}, selective
computation aims to preserve generation
quality~\citep{liu2026region,zhang2025trainingfree}.
ToCa uses token importance for caching~\citep{zou2025accelerating},
while WorldCache exploits trajectory predictability for reuse
and extrapolation~\citep{feng2026worldcache}.
However, observed-input relevance and visual fidelity do not
directly determine which evolving future regions support
action prediction.
Our method therefore uses action-to-future attention to guide
future-frame token selection and reuses selected positions across
denoising steps.

\section{Preliminaries and Motivation}

\subsection{World Action Models}

As shown in \autoref{fig:comparison}, WAMs with a shared
Transformer sequence jointly denoise future-frame tokens
$\mathbf{z}_{\mathrm v}$ and an $H$-step action chunk
$\mathbf{a}_{1:H}$~\citep{ye2026dreamzero,agarwal2026cosmos3}.
Conditioned on the current observation, task instruction,
and robot state, successive denoising steps refine the
same predicted future and action chunk.
During this process, action tokens attend to future-frame
tokens, allowing information from the predicted future
to inform action updates.
Reducing future-frame computation therefore calls for
understanding how action tokens access this information.

\paragraph{Action-to-Future Attention.}
Attention from action queries to future-frame keys
provides a token-level view of this interaction.
For a predicted future comprising $F$ latent frames
with $N_s$ tokens per frame, let $\mathbf{z}_{f,j}$
denote the future-frame token at spatial position $j$
in frame $f$, and $\mathbf{a}_i$ the $i$-th action token.
At Transformer layer $\ell$ and denoising step $\tau$
(starting from $\tau=0$),
let $A_{\ell,h}^{(\tau)}(i,f,j)$ denote the attention
weight from the query of action token $\mathbf{a}_i$
to the key of future-frame token $\mathbf{z}_{f,j}$
in head $h$.
These weights retain the normalization over all keys
visible to each action query.
We define action-to-future attention by averaging
these weights across heads:
\begin{equation}
U_{\ell}^{(\tau)}(i,f,j)
=
\frac{1}{N_h}
\sum_{h=1}^{N_h}
A_{\ell,h}^{(\tau)}(i,f,j),
\label{eq:action-to-future-attention}
\end{equation}
where $N_h$ is the number of attention heads,
$i=1,\ldots,H$ indexes action tokens,
$f=1,\ldots,F$ indexes future frames, and
$j=1,\ldots,N_s$ indexes spatial tokens within each frame.
A larger value indicates stronger attention from action token $i$ to future-frame token $(f,j)$. We use this attention as a low-cost proxy for token relevance when allocating sparse computation.
We measure cross-frame and cross-step overlap by summing
the smaller of the two normalized attention values at each
spatial position (see Appendix~\ref{app:audit-overlap}).

\subsection{Observations}\label{sec:observations}
This section examines how action-to-future attention
is distributed across future frames and spatial
positions, and how these patterns vary across
Transformer layers and denoising steps.
We conduct this analysis on Cosmos3-Edge and
Cosmos3-Nano using RoboLab~\citep{yang2026robolab}.

\begin{figure}[t]
  \centering
  \includegraphics[width=\textwidth]{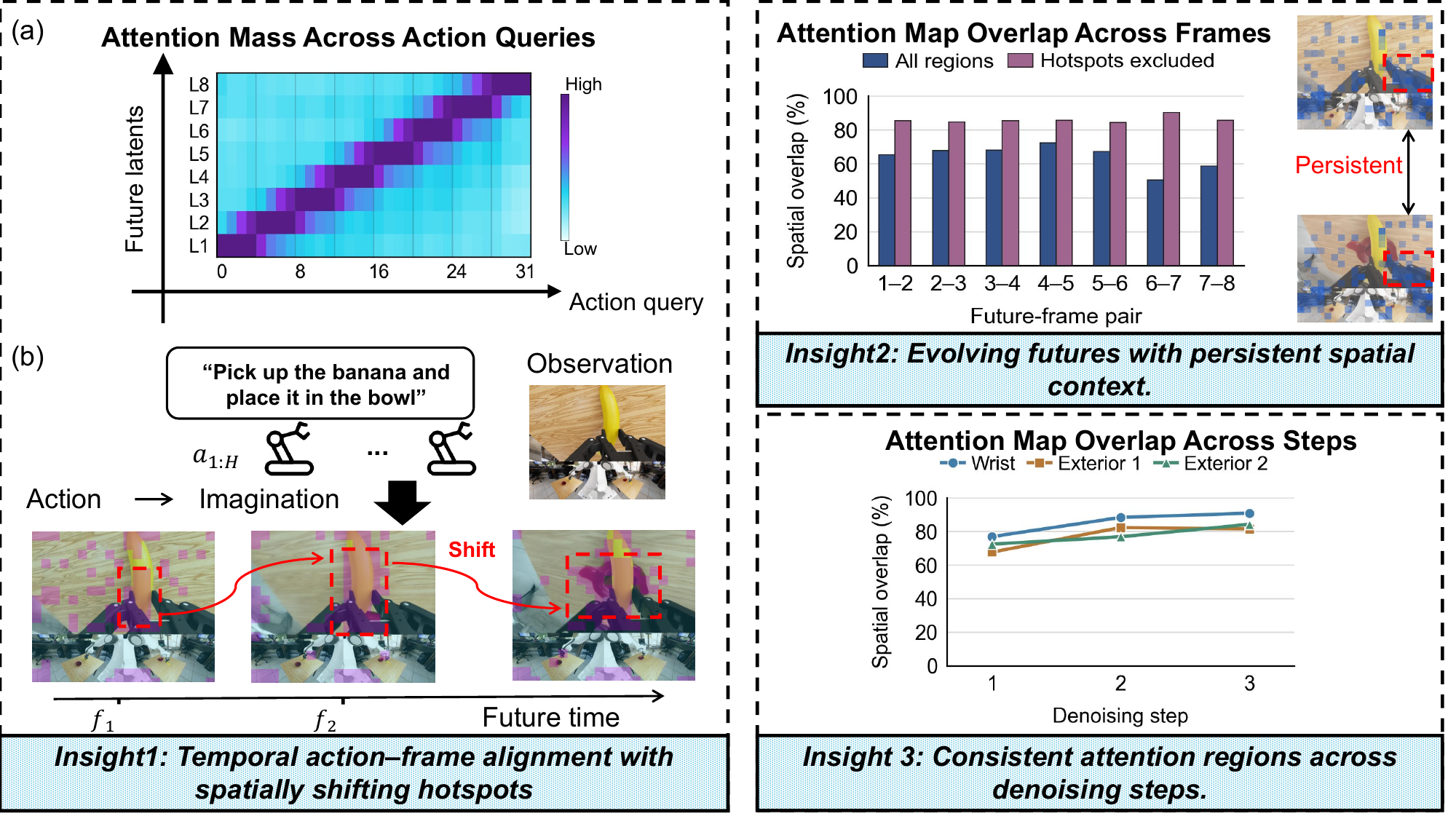}
  \caption{Insight 1: (a) Action queries exhibit temporal alignment
        with future frames, with attention extending to
        neighboring frames near temporal group boundaries.
        (b) Action-relevant hotspots shift spatially across
        future frames.
        Insight 2: Excluding hotspots yields higher attention overlap between adjacent frames.
        Insight 3: Attention maps exhibit substantial spatial
        overlap between consecutive denoising steps in all three camera views.}
  \label{fig:observation}
\end{figure}

\paragraph{Action--Frame Alignment and Spatial Variation.}
We identify two complementary patterns in action-to-future
attention.
First, \autoref{fig:observation} (Insight 1(a)) shows an
approximately diagonal band in the frame-level attention map:
each future frame predominantly receives attention from a
contiguous group of roughly $H/F$ action queries.
Queries near group boundaries also attend to neighboring
frames, indicating that the alignment extends across temporal
group boundaries.
Second, \autoref{fig:observation} (Insight 1(b)) shows that
attention hotspots shift spatially across future frames.
These patterns motivate scoring each future frame using its
temporally associated action queries and selecting spatial
regions separately across frames.

\paragraph{Cross-Frame Context Consistency.}
We further compare cross-frame attention overlap with and
without hotspot regions.
As shown in \autoref{fig:observation} (Insight 2),
overlap increases after hotspot exclusion, indicating greater
consistency in the spatial distribution of the remaining attention.
The visualizations also show attention to contextual regions,
including parts of the gripper outside the object-contact area.
Together, these observations suggest that persistent spatial
context complements frame-specific hotspots, motivating shared
anchor positions alongside per-frame token selection.

\paragraph{Cross-Step Attention Consistency.}
We next examine how spatial attention distributions change
across denoising steps that refine the same predicted future.
As shown in \autoref{fig:observation} (Insight 3),
the distributions exhibit substantial overlap between consecutive
steps in each of the three camera views.
Thus, future representations continue to evolve while their
spatial attention distributions remain relatively consistent.
This observation motivates reusing token selections across
denoising steps to reduce repeated selection overhead. Appendix~\ref{app:audit-overlap} defines the overlap measure
and provides cross-step and cross-frame comparisons
with attention visualizations.

\section{Methodology}
\label{sec:method}

\begin{figure}[t]
    \centering
    \includegraphics[width=\linewidth]{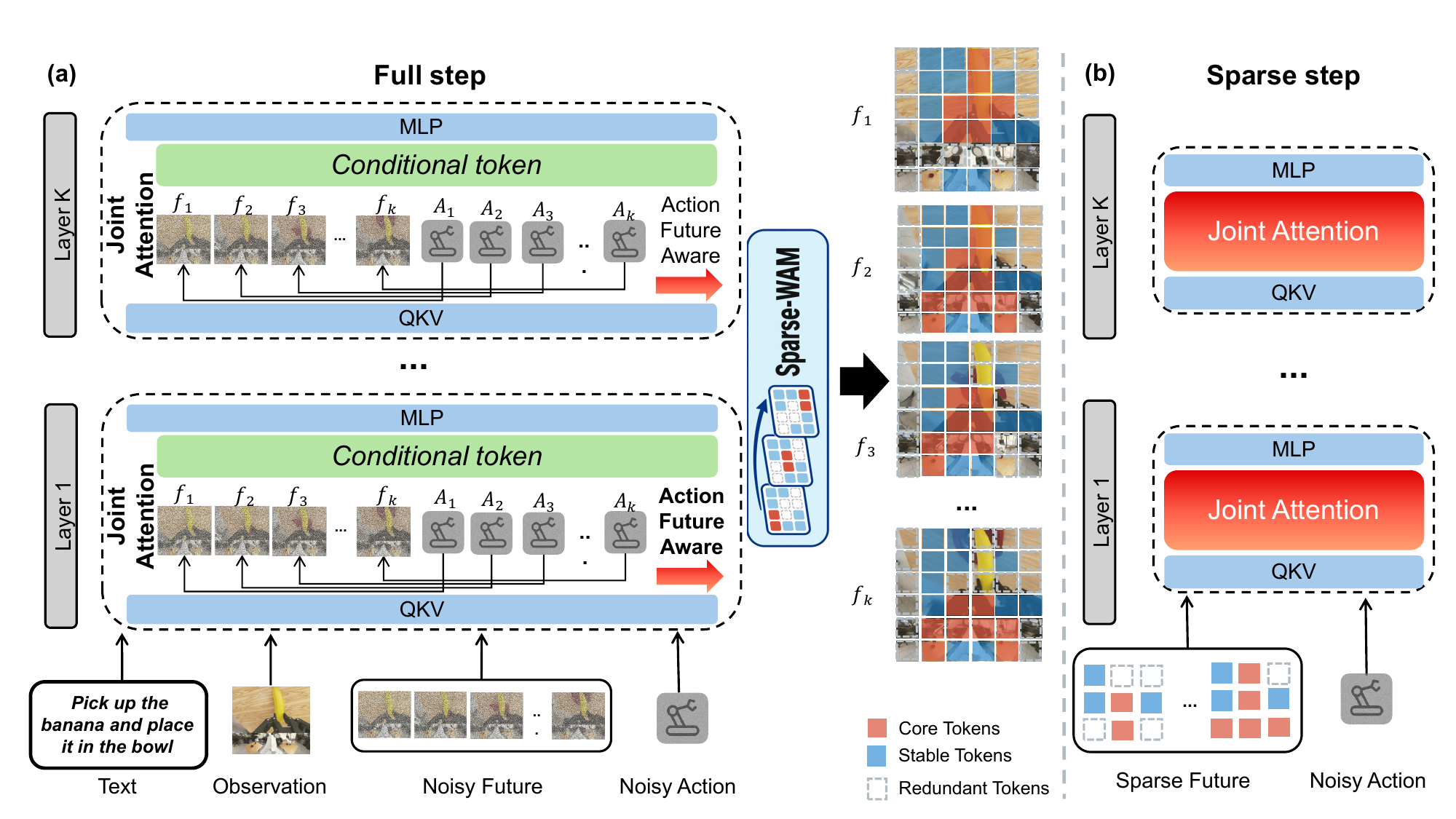}
    \caption{Overview of \method{}.
    Future-frame token positions are selected using action-to-future
    attention and reused during sparse denoising.}
    \label{fig:overview}
\end{figure}

As illustrated in \autoref{fig:overview}, \method{} uses
action-to-future attention to allocate computation within
imagined futures.
In \autoref{subsec:mask-construction}, we combine frame-specific
core tokens with shared spatial anchors to retain changing
attention hotspots and persistent context.
In \autoref{subsec:efficient-execution}, we introduce
\textbf{Pilot}, an engine that combines lightweight attention
profiling with cross-step reuse to execute these selections
efficiently.

\subsection{Online Action-Guided Token Selection}
\label{subsec:mask-construction}

As discussed in \autoref{fig:observation},
the token selection needs to follow the hotspots
of each future frame while retaining contextual regions
shared across frames.
We address these complementary needs with $K_c$ frame-specific
core tokens and $K_s$ shared spatial anchors per frame,
where $K_c+K_s\leq N_s$.
Selection uses attention from the dense conditional forward
at a full denoising step $\tau_d$.
For brevity, we write
$U_\ell(i,f,j)=U_\ell^{(\tau_d)}(i,f,j)$ below.

\paragraph{Scoring Action-Relevant Regions.}
The temporal alignment in \autoref{fig:observation}
(Insight~1(a)) suggests that each future frame should be
scored using its associated action queries.
Assuming that $H$ is divisible by $F$, we partition the
$H$ queries into $F$ consecutive groups of size $H/F$,
denoting the group for frame $f$ by $\mathcal{I}_f$.
The spatial score at layer $\ell$ is
\begin{equation}
    S_\ell(f,j)
    =
    \sum_{i\in\mathcal{I}_f}
    \alpha_{f,i}U_\ell(i,f,j),
    \label{eq:frame-aligned-score}
\end{equation}
where the nonnegative weights $\alpha_{f,i}$ sum to one.
Central queries receive larger weights because queries near
group boundaries also attend to neighboring frames.

To identify layers whose attention provides localized signals for token selection, we assign each layer a score $Q_\ell=R_\ell(1-E_\ell)$.
Here, $R_\ell$ measures frame-aligned attention mass,
and $E_\ell$ is the normalized spatial entropy averaged
across frames~\citep{zhang2025attention}.
We select the $K_{\mathrm{layer}}$ highest-scoring layers,
forming $\mathcal{L}_{\mathrm{core}}$, and aggregate their
spatial scores:
\begin{equation}
    V(f,j)
    =
    \frac{
        \sum_{\ell\in\mathcal{L}_{\mathrm{core}}}
        Q_\ell S_\ell(f,j)
    }{
        \sum_{\ell\in\mathcal{L}_{\mathrm{core}}}Q_\ell
    }.
    \label{eq:core-score}
\end{equation}
Appendix~\ref{app:selection-details} provides the query-group
definition and layer-quality calculations.

\paragraph{Frame-Specific Core Tokens.}
Since attention hotspots shift across future frames,
each frame selects its own core positions using $V(f,j)$.
We first identify positions with high core scores in at least
one future frame, then select frame-specific tokens from
this common candidate pool.

Let $\mathcal{J}=\{1,\ldots,N_s\}$ denote all spatial positions.
The candidate pool contains $N_s-K_s$ positions, preserving
capacity for the shared anchors.
We construct the pool and select core positions as
\begin{align}
    \mathcal{P}
    &=
    \operatorname{TopK}_{j\in\mathcal{J}}
    \left(
        \max_f V(f,j),\,N_s-K_s
    \right),
    \label{eq:core-candidate-pool}
    \\
    \mathcal{C}_f
    &=
    \operatorname{TopK}_{j\in\mathcal{P}}
    \left(
        V(f,j),K_c
    \right),
    \label{eq:core-selection}
\end{align}
where $\operatorname{TopK}$ returns the indices of the
largest scores.
The maximum across frames determines the common candidate
pool, while each frame's own scores determine its core
selection, allowing the retained positions to follow
spatially shifting hotspots.

\paragraph{Shared Spatial Anchor Tokens.}
The increased cross-frame attention overlap after hotspot
exclusion (\autoref{fig:observation}, Insight~2) suggests
that contextual regions receive more consistent attention
across future frames.
We therefore complement frame-specific core tokens
with shared spatial anchors.

To avoid duplicating core positions, anchors are selected
from $\mathcal{E}$, the positions not selected as core
tokens in any frame.
Because all core selections lie within the common pool
$\mathcal{P}$ of size $N_s-K_s$, at least $K_s$ positions
remain available for anchors.
Appendix~\ref{app:selection-details} provides the formal
candidate-set definition and availability guarantee.

To rank these candidates, we aggregate $S_\ell(f,j)$
over all layers using the quality weights $Q_\ell$,
including signals beyond the layers selected for
core localization.
Let $\mu_j$ and $\mathrm{CV}_j$ denote the cross-frame
mean and coefficient of variation of this aggregated
score at position $j$.
We select positions with strong and consistent attention:
\begin{equation}
    \mathcal{S}
    =
    \operatorname{TopK}_{j\in\mathcal{E}}
    \left(
        \frac{\mu_j}{1+\mathrm{CV}_j},
        K_s
    \right).
    \label{eq:stable-selection}
\end{equation}
The numerator rewards attention strength, while
the denominator penalizes cross-frame variation.
Each frame retains the positions $\mathcal{C}_f\cup\mathcal{S}$.
The two sets are disjoint, giving exactly $K_c+K_s$ tokens
per frame and a fixed sequence length for sparse execution.
Only anchor positions are shared: their representations
remain frame-specific and are updated separately. Appendix~\ref{app:audit-visualization} visualizes the selected
core tokens and shared anchors.

\subsection{Pilot: Online Selection and Sparse Execution}
\label{subsec:efficient-execution}

In this section, we implement sparse execution through lightweight
attention profiling, cross-step selection reuse, and cached
prediction updates.
Together, these mechanisms reduce selection and execution
overhead while maintaining the sampling updates required
by joint visual--action denoising.

\paragraph{Low-Overhead Attention Profiling.}
At each full step, Pilot obtains the required scores from
the dense conditional forward without an additional
network forward.
The scorer computes only the query--key products between
each future frame and its associated action-query group.
It reuses the original log-sum-exp normalizers computed over
all visible keys, preserving the attention mass used
in layer-quality scoring.
The resulting probabilities are aggregated directly into
$S_\ell(f,j)$, avoiding materialization of the full attention
matrix or an additional value-weighted attention output.
Appendix~\ref{app:profiling-details} provides the extraction
formula and implementation details.

\paragraph{Cross-Step Reuse and Compact Execution.}
The observed cross-step attention consistency motivates
reusing selections within each action chunk.
A full step constructs the selection and caches the visual
velocity predictions; subsequent sparse steps reuse both.
Given the short denoising schedules of the evaluated WAMs,
our default configuration uses only the initial step for
full computation.
Selections and cached predictions are recomputed for each
new action chunk.
Alternative refresh schedules are evaluated in
Appendix~\ref{app:audit-selection-reuse}.

During sparse steps, the selected future tokens are processed
as a compact sequence together with all observation and action
tokens.
Pilot reuses their original-position indices and compatible
packing metadata across Transformer layers and denoising steps.
The fixed per-frame token budget maintains a constant sequence
length despite different spatial selections across frames.

\paragraph{Sampling Updates for Omitted Regions.}
Future tokens omitted from Transformer computation still
participate in the sampling process.
Let $\tau_d$ denote the most recent full step and
$\mathbf{M}_{\mathrm v}$ the retained-position mask mapped
to the full future-latent grid.
Pilot combines current predictions at retained positions
with cached predictions elsewhere:
\begin{equation}
    \widetilde{\mathbf v}_{\mathrm v}^{(\tau)}
    =
    \mathbf M_{\mathrm v}\odot
    \mathbf v_{\mathrm v}^{(\tau)}
    +
    (\mathbf 1-\mathbf M_{\mathrm v})\odot
    \mathbf v_{\mathrm v}^{(\tau_d)},
    \qquad \tau>\tau_d,
    \label{eq:cached-visual-prediction}
\end{equation}
where $\odot$ denotes element-wise multiplication,
$\mathbf v_{\mathrm v}^{(\tau)}$ contains current predictions
scattered to their retained positions in the full grid,
and $\mathbf v_{\mathrm v}^{(\tau_d)}$ is the cached prediction
from the full step.
The original sampler uses
$\widetilde{\mathbf v}_{\mathrm v}^{(\tau)}$
to update all future latents, including omitted regions,
while action predictions are recomputed at every step.

\section{Experiments}
\label{sec:experiments}

\subsection{Experimental Settings}
\label{subsec:experimental-setting}

\paragraph{Benchmarks and Models.}
We evaluate \method{} using FastWAM-Joint~\citep{yuan2026fastwam}
on LIBERO~\citep{liu2023libero} and
LIBERO-Plus~\citep{fei25libero-plus}, and Cosmos 3 Nano Policy
and Cosmos 3 Edge~\citep{agarwal2026cosmos3}
on RoboLab-120~\citep{yang2026robolab}.
LIBERO comprises four manipulation suites, LIBERO-Plus adds
seven perturbation categories, and RoboLab-120 contains
120 tasks across three difficulty levels.
We also evaluate FastWAM-Joint on real-world object packing,
cup stacking, and battery insertion tasks.
Figure~\ref{fig:tasks} illustrates representative tasks.
Appendix~\ref{lab: benchmarks} provides benchmark protocols.

\paragraph{Baselines and Evaluation Setup.}
We compare against \swamrev{five} inference acceleration methods:
ToCa~\citep{zou2025accelerating},
WorldCache~\citep{feng2026worldcache},
SpecPrune-VLA~\citep{wang2026specprune},
C\textsuperscript{3}ache~\citep{zhao2026c3ache}, and
\swamrev{ActionCache~\citep{oi2026actioncache}}.
\swamrev{We measure inference latency for all policies using CUDA events
on an NVIDIA RTX 4090 and enable the same generic backend optimizations
used by Pilot for all compared acceleration methods.}
Appendix~\ref{app:audit-implementation} provides implementation
details, while Appendix~\ref{app:audit-baselines}
describes the baseline adaptations.

\paragraph{Evaluation Metrics.}
We report task success rate (SR), inference speedup, and
FLOPs as a percentage of the corresponding dense model.
\swamrev{The main LIBERO and RoboLab-120 results report speedups
  relative to dense eager inference.}
Appendix~\ref{app:audit-timing} details the timing protocol
and reports on an execution backend ablation and comparisons
with matched execution backends.

\subsection{Simulation Results}
\label{subsec:simulation-results}

\begin{table}[t]
    \centering
    \caption{
        Performance and inference efficiency on LIBERO.
    }
    \label{tab:libero-results}
    \small
    \setlength{\tabcolsep}{4pt}
    \renewcommand{\arraystretch}{1.1}

    \begin{tabularx}{\linewidth}{
        @{}
        p{0.28\linewidth}
        *{4}{>{\centering\arraybackslash}X}
        *{3}{>{\centering\arraybackslash}p{0.10\linewidth}}
        @{}
    }
        \toprule
        \multirow{2}{*}{Method}
        & \multicolumn{4}{c}{Success Rate (\%)}
        & \multirow{2}{*}{\shortstack{Avg.\\SR (\%)}}
        & \multirow{2}{*}{\shortstack{Avg.\\Speedup}}
        & \multirow{2}{*}{FLOPs} \\
        \cmidrule(lr){2-5}
        & Spatial & Object & Goal & Long
        & & & \\
        \midrule

        FastWAM-Joint
        & 99.00\% & 100.00\% & 98.20\% & 97.80\%
        & 98.75\% & $1.00\times$ & 100.00\% \\

        \swamrev{+ ActionCache}
        & \swamrev{99.20\%} & \swamrev{98.40\%}
        & \swamrev{96.80\%} & \swamrev{95.60\%}
        & \swamrev{97.50\%} & \swamrev{$1.33\times$}
        & \swamrev{74.93\%} \\

        + ToCa
        & 98.80\% & 99.80\% & 98.60\% & 98.60\%
        & 98.95\% & $1.09\times$ & 69.19\% \\

        + C\textsuperscript{3}ache
        & 98.60\% & 99.40\% & 97.60\% & 97.40\%
        & 98.25\% & $1.30\times$ & 75.00\% \\

        + SpecPrune-VLA
        & 97.20\% & 99.80\% & 96.80\%
        & 92.40\% & 96.55\% & $1.43\times$ & 73.53\% \\

        + WorldCache
        & 99.60\% & 99.80\% & 98.00\% & 98.60\%
        & 99.00\% & $1.62\times$ & 60.00\% \\

        + \textbf{\method{}}
        & 98.80\% & 99.20\% & 98.60\% & 97.20\%
        & 98.45\% & \textbf{$1.98\times$} & 49.65\% \\

        \bottomrule
    \end{tabularx}
\end{table}

\begin{table}[t]
    \centering
    \caption{
        Performance and inference efficiency on RoboLab-120.
    }
    \label{tab:robolab-results}
    \small
    \setlength{\tabcolsep}{4pt}
    \renewcommand{\arraystretch}{1.1}

    \begin{tabularx}{\linewidth}{
        @{}
        p{0.28\linewidth}
        *{3}{>{\centering\arraybackslash}X}
        *{3}{>{\centering\arraybackslash}p{0.10\linewidth}}
        @{}
    }
        \toprule
        \multirow{2}{*}{Method}
        & \multicolumn{3}{c}{Success Rate (\%)}
        & \multirow{2}{*}{\shortstack{Avg.\\SR (\%)}}
        & \multirow{2}{*}{\shortstack{Avg.\\Speedup}}
        & \multirow{2}{*}{FLOPs} \\
        \cmidrule(lr){2-4}
        & Simple & Moderate & Complex
        & & & \\
        \midrule

        Cosmos3-Edge
        & 25.60\% & 23.30\% & 11.80\%
        & 22.90\% & $1.00\times$ & 100.00\% \\
        
        \swamrev{+ ActionCache}
        & \swamrev{24.84\%} & \swamrev{23.59\%} & \swamrev{10.59\%}
        & \swamrev{22.42\%} & \swamrev{$1.22\times$} & \swamrev{97.82\%} \\

        + ToCa
        & 9.84\% & 14.10\% & 1.18\%
        & 10.00\% & $1.26\times$ & 65.68\% \\

        + C\textsuperscript{3}ache
        & 9.06\% & 12.82\% & 0.59\%
        & 9.08\% & $1.51\times$ & 75.14\% \\

        + SpecPrune-VLA
        & 10.31\% & 14.10\% & 0.00\%
        & 10.08\% & $1.52\times$ & 50.35\% \\

        + WorldCache
        & 21.56\% & 21.03\% & 11.18\%
        & 19.92\% & $1.52\times$ & 75.14\% \\

        + \textbf{\method{}}
        & 25.00\% & 25.90\% & 8.82\%
        & 23.00\% & \textbf{$1.85\times$} & 60.73\% \\

        \midrule

        Cosmos3-Nano-Policy
        & 40.63\% & 35.38\% & 25.29\%
        & 36.75\% & $1.00\times$ & 100.00\% \\
        \swamrev{+ ActionCache}
        & \swamrev{40.16\%} & \swamrev{32.05\%} & \swamrev{21.18\%}
        & \swamrev{34.83\%} & \swamrev{$1.13\times$} & \swamrev{98.75\%} \\

        + ToCa
        & 9.69\% & 12.82\% & 0.00\%
        & 9.33\% & $1.33\times$ & 65.21\% \\

        + C\textsuperscript{3}ache
        & 16.09\% & 13.08\% & 1.18\%
        & 13.00\% & $1.48\times$ & 75.00\% \\

        + SpecPrune-VLA
        & 27.50\% & 23.08\% & 11.76\%
        & 23.83\% & $1.88\times$ & 54.79\% \\

        + WorldCache
        & 36.56\% & 30.00\% & 20.00\%
        & 32.08\% & $1.48\times$ & 75.00\% \\

        + \textbf{\method{}}
        & 40.16\% & 32.82\% & 24.12\%
        & 35.50\% & \textbf{$1.81\times$} & 59.78\% \\

        \bottomrule
    \end{tabularx}
\end{table}

\paragraph{Results on LIBERO.}
Table~\ref{tab:libero-results} reports performance and
inference efficiency across the four LIBERO suites.
Speedups in both main tables are measured relative to dense
eager inference and reflect the combined effects of each
acceleration method and execution optimizations.
\method{} achieves the largest speedup among the evaluated
methods ($1.98\times$), reducing FLOPs by 50.35\% while
attaining 98.45\% success, 0.30 percentage points below
dense inference.
ToCa, WorldCache, and C\textsuperscript{3}ache also maintain
high success rates but achieve smaller speedups, up to
$1.62\times$.
SpecPrune-VLA incurs a larger performance drop, particularly
on long-horizon tasks.

\paragraph{Results on RoboLab-120.}
Table~\ref{tab:robolab-results} shows that \method{} achieves
the highest average success rates among the accelerated
variants on both Cosmos backbones: 23.00\% on Edge and
35.50\% on Nano, compared with 22.90\% and 36.75\% for
dense inference.
With execution optimizations enabled, it achieves respective
speedups of $1.85\times$ and $1.81\times$ over dense eager
inference.
WorldCache preserves performance better than ToCa and
C\textsuperscript{3}ache but is slower than \method{}.
SpecPrune-VLA is slightly faster on Nano ($1.88\times$),
at a success rate 11.67 percentage points below ours.
On Edge, \method{} outpaces our SpecPrune-VLA adaptation
($1.85\times$ versus $1.52\times$) despite using more FLOPs
(60.73\% versus 50.35\% of dense inference), showing that
FLOPs alone do not determine latency.
Our implementation reuses token selections and maintains
fixed sparse sequence lengths to support compiled execution.

\begin{swamrevision}
    \paragraph{Results on LIBERO-Plus.}
    On LIBERO-Plus shown in Figure~\ref{fig:audit-perturbations}, FastWAM skips iterative video denoising and achieves a high speedup over dense eager FastWAM-Joint
    (Table~\ref{tab:libero-plus-results}). Its success rates under camera
    and sensor-noise perturbations drop sharply. Sparse-WAM selectively retains future tokens during denoising and achieves \(70.00\%\) mean success, with a \(1.95\times\) speedup over dense eager inference. Appendix~\ref{app:audit-implementation} provides configuration details and Table~\ref{tab:swam-rev-robustness} provides further experiments.
\end{swamrevision}

\begin{table}[t]
    \begin{swamrevision}
    \centering
    \caption{
        Success rates (\%) and inference efficiency
        on the evaluated LIBERO-Plus.
    }
    \label{tab:libero-plus-results}
    \footnotesize
    \setlength{\tabcolsep}{2pt}
    \renewcommand{\arraystretch}{1.1}

    \begin{tabularx}{\linewidth}{
        @{}
        p{0.22\linewidth}
        *{8}{>{\centering\arraybackslash}X}
        c
        c
        @{}
    }
        \toprule
        \multirow{2}{*}{Method}
        & \multicolumn{8}{c}{Success Rate (\%)}
        & \multirow{2}{*}{Latency (ms)}
        & \multirow{2}{*}{Speedup}
        \\
        \cmidrule(lr){2-9}
        & Camera & Robot & Lang. & Light
        & Backg. & Noise & Layout & Avg.
        & & \\
        \midrule

        FastWAM
        & 26.25 & 41.25 & 63.75 & 77.50
        & 57.50 & 36.25 & 61.25 & 51.96
        & 90.1  & \(4.79\times\)\\

        FastWAM-Joint (dense)
        & \textbf{51.25} & 51.25 & 87.50 & 95.00
        & 66.25 & 61.25 & 83.75 & 70.89
        & 431.7 & \(1.00\times\)\\

        \quad + Sparse-WAM
        & 46.25 & 52.50 & 90.00 & 92.50
        & 62.50 & 62.50 & 83.75 & 70.00
        & 221.0 & \(1.95\times\)  \\

        \bottomrule
    \end{tabularx}\par
\end{swamrevision}
\end{table}

\subsection{Real-World Experiments}
\label{subsec:real-world}

\begin{table}[t]
    \centering
    \caption{
        Real-world manipulation performance on AgileX Cobot Magic.
    }
    \label{tab:real-world-results}
    \small
    \setlength{\tabcolsep}{6pt}
    \begin{tabular}{lccccrr}
        \toprule
        \multirow{2}{*}{Method}
        & \multicolumn{4}{c}{Success Rate (\%)}
        & \multirow{2}{*}{\shortstack{Latency\\(ms)}}
        & \multirow{2}{*}{Speedup} \\
        \cmidrule(lr){2-5}
        & Pack objects
        & Stack cups
        & Battery assembly
        & Average
        & & \\
        \midrule
        FastWAM-Joint
        & 75.00 & 75.00 & 83.33 & 77.78
        & 501 & $1.00\times$ \\
        \quad + \textbf{\method{}}
        & 83.33 & 75.00 & 66.67 & 75.00
        & 242 & $2.08\times$ \\
        \bottomrule
    \end{tabular}
\end{table}
\vspace{-8pt}
\begin{figure}[t]
    \centering
    \includegraphics[width=\linewidth]{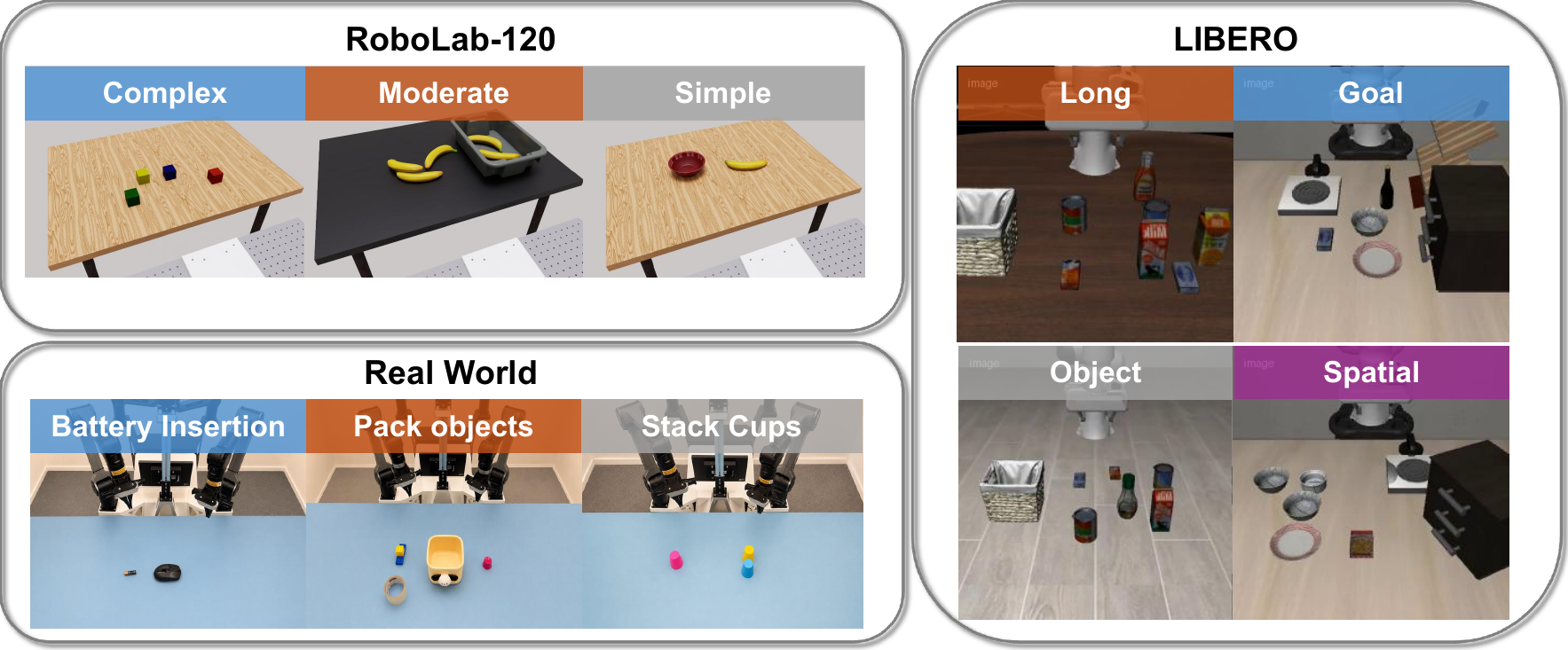}
    \caption{Tasks on LIBERO, RoboLab-120 and Real World.}
    \label{fig:tasks}
\end{figure}

We evaluate \method{} on the AgileX Cobot Magic platform,
which is equipped with three cameras providing different
viewpoints: one primary camera and two wrist-mounted cameras.
We fine-tune FastWAM-Joint using our collected data,
following the configuration detailed in
Appendix~\ref{app:fine-tune}.

We design three tasks: object packing,
cup stacking, and battery insertion, as shown in
Figure~\ref{fig:tasks}.
Table~\ref{tab:real-world-results} reports the real-world
performance of \method{}.
Our method achieves a $2.08\times$ speedup with an average
success rate of 75.00\%, compared with 77.78\% for dense
FastWAM-Joint.
These results demonstrate the potential of \method{} to
accelerate real-world WAM inference by reducing redundant
computation over imagination tokens while largely
preserving average task success.

\subsection{Ablation Studies}
\label{subsec:ablation}

\begin{figure}[H]
    \centering
    \begin{subfigure}[t]{0.48\linewidth}
        \centering
        \vspace{0pt}
        \includegraphics[width=\linewidth]
        {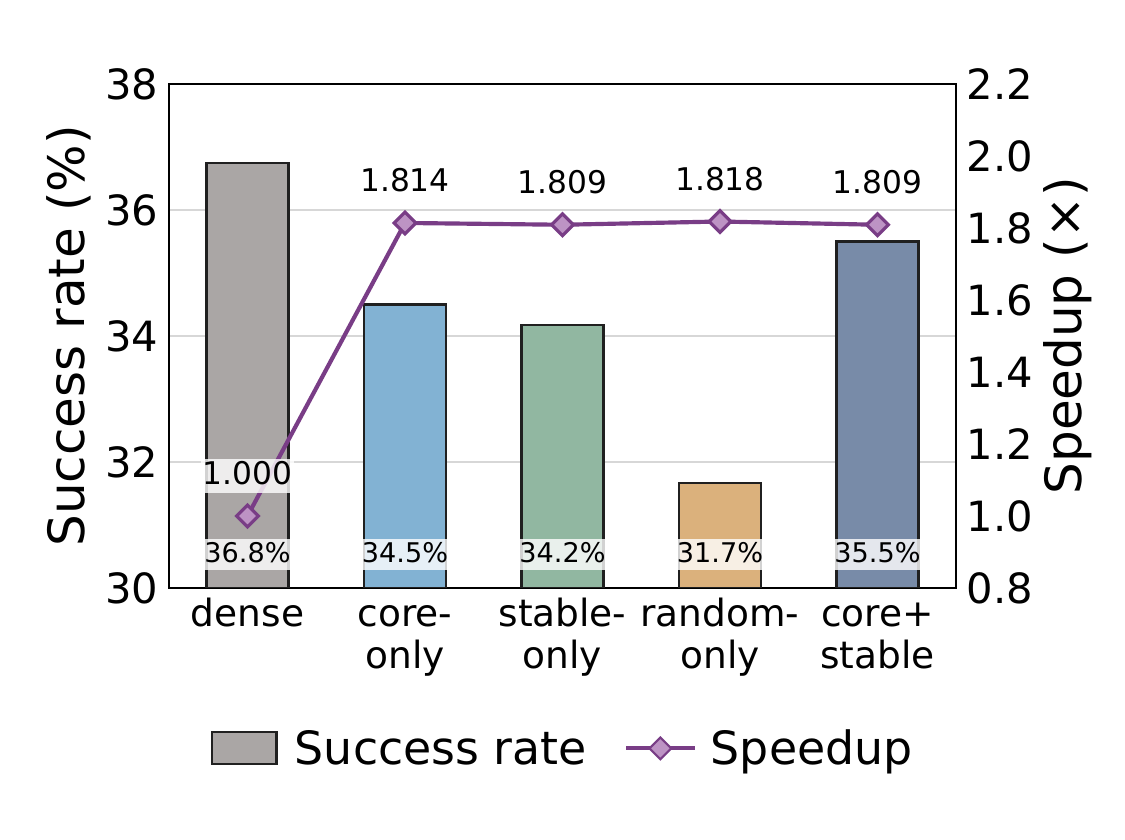}
        \vspace{-15pt}
        \caption{Core tokens and stable anchors.}
        \label{fig:core-stable-ablation}
    \end{subfigure}
    \hfill
    \begin{subfigure}[t]{0.48\linewidth}
        \centering
        \vspace{0pt}
        \includegraphics[width=\linewidth]
        {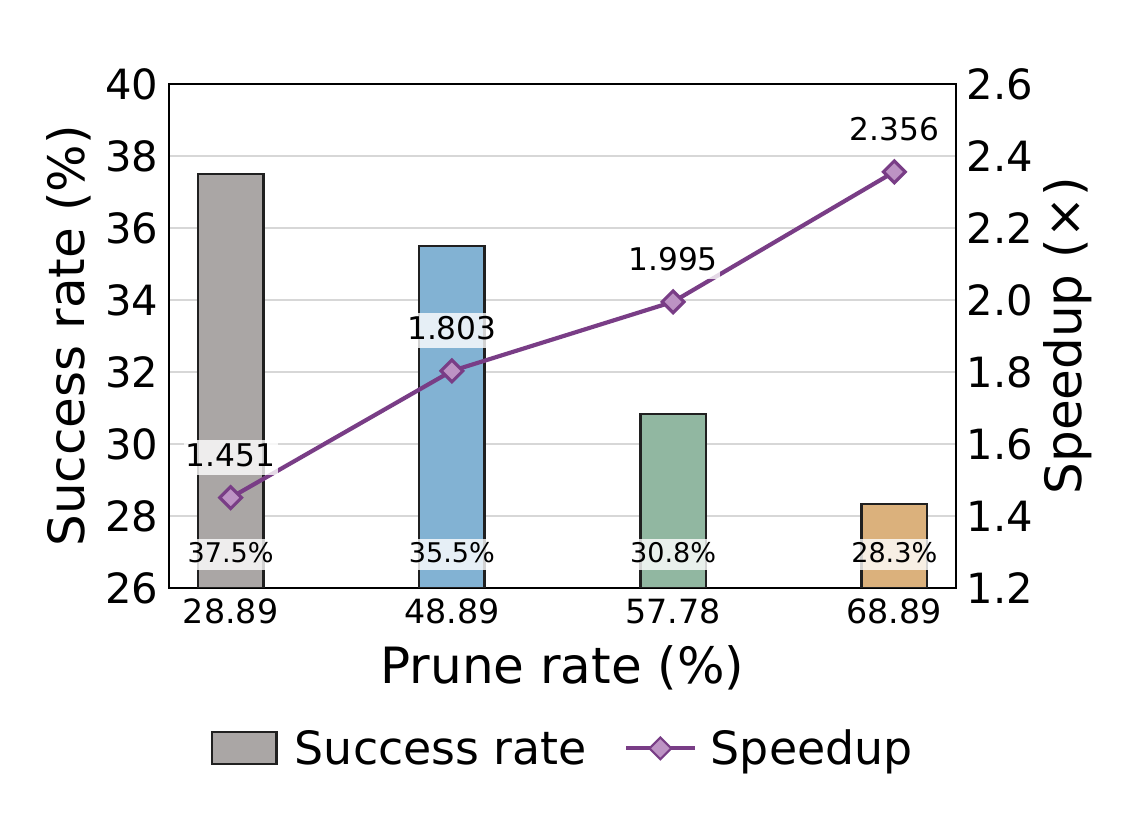}
        \vspace{-15pt}
        \caption{Future-token pruning ratio.}
        \label{fig:token-budget-ablation}
    \end{subfigure}
    \vspace{-5pt}
    \caption{
        Ablation studies of token selection components
        and future-token pruning ratios.
    }
    \label{fig:ablation-studies}
\end{figure}

\paragraph{Component Ablation.}
Figure~\ref{fig:core-stable-ablation} compares token
selection strategies on Cosmos 3 Nano Policy,
with all sparse variants retaining 184 of 360 tokens
per future frame.
Combining $K_c=80$ core tokens with $K_s=104$ stable
anchors achieves a 35.5\% success rate, outperforming
both core-only and stable-only selection.
Random selection achieves only 31.7\%, compared
with 36.8\% under dense inference.
The combined variant maintains approximately
$1.81\times$ speedup, comparable to core-only selection.
These results support the complementary roles of
frame-specific core tokens and shared stable anchors
in preserving action-relevant imagination,
with little additional inference overhead. \swamrev{Table~\ref{tab:fastwam-components} reports the corresponding
component ablation on FastWAM-Joint.}

\paragraph{Pruning Ratio.}
Figure~\ref{fig:token-budget-ablation} examines the
effect of future-token pruning on Cosmos 3 Nano Policy. We define the future-token pruning ratio as
$r=1-K_{\mathrm{ret}}/N_s$, where $K_{\mathrm{ret}}$
is the number of retained future tokens per frame.
Increasing the pruning ratio from 28.89\% to 68.89\%
improves speedup from $1.45\times$ to $2.36\times$,
while success rate decreases from 37.5\% to 28.3\%.
These results illustrate the efficiency--control
trade-off: more aggressive pruning accelerates
inference at the expense of task success. \swamrev{Appendix~\ref{app:audit-additional-ablations} further ablates
informative-layer aggregation and temporal action weighting and Table~\ref{tab:fastwam-pruning} reports the pruning-ratio sweep
on FastWAM-Joint.}

\begin{swamrevision}
\paragraph{Pilot and Backend Speedups.}
Sparse-WAM uses a fixed token budget to keep execution shapes static
and LSE-based attention ranking to reduce token-selection overhead.
Under the same backend optimizations, Sparse-WAM achieves the highest
speedup among the evaluated acceleration methods on LIBERO and
Cosmos3-Edge, and the second-highest on Cosmos3-Nano-Policy
(Tables~\ref{tab:libero-results} and~\ref{tab:robolab-results}).
To quantify the gains beyond generic
backend optimizations, we compare Sparse-WAM with Pilot against dense
inference under the same \texttt{torch.compile} and CUDA Graph settings.
The full Pilot configuration combines \texttt{torch.compile} and CUDA Graphs with LSE-based attention scoring and cross-step caching of token selections and visual predictions.
Sparse-WAM with Pilot achieves \(1.56\text{--}1.59\times\) speedups
across all three backbones. Appendix~\ref{app:audit-timing} reports
the full eager and optimized timing breakdown.

\end{swamrevision}

\begin{table}[!htbp]
  \begin{swamrevision}
    \centering
    \caption{Speedups of Sparse-WAM with Pilot.}
    \small
    \begin{tabular}{@{}lrr@{}}
      \toprule
      Backbone & vs.\ dense eager & vs.\ optimized dense \\
      \midrule
      FastWAM-Joint       & \(1.98\times\) & \(\mathbf{1.58\times}\) \\
      Cosmos3-Nano-Policy & \(1.81\times\) & \(\mathbf{1.59\times}\) \\
      Cosmos3-Edge        & \(1.85\times\) & \(\mathbf{1.56\times}\) \\
      \bottomrule
    \end{tabular}\par
  \end{swamrevision}
\end{table}

\begin{swamrevision}
\paragraph{LIBERO-Plus Robustness Ablation.}
With a small token budget, even modest fluctuations in attention rankings
can push relevant future tokens below the selection cutoff. When full
refreshes are infrequent, these selection errors and stale cached
predictions can persist across denoising steps. Retaining more tokens
improves mean success, while more frequent full refreshes update both
token selections and cached predictions and improve success under camera
perturbations. At \(42.9\%\) pruning, Sparse-WAM preserves a success rate close to the baseline with a \(1.94\times\) speedup.
Speedups in Table~\ref{tab:swam-rev-robustness} are relative to dense eager inference.
\par
\end{swamrevision}

\begin{table}[!htbp]
  \begin{swamrevision}
    \centering
    \caption{LIBERO-Plus robustness ablations with FastWAM-Joint.}
    \label{tab:swam-rev-robustness}
    \footnotesize
    \setlength{\tabcolsep}{2.5pt}
    \begin{tabular}{@{}rl*{10}{r}@{}}
      \toprule
      & & \multicolumn{8}{c}{Success rate (\%)} & & \\
      \cmidrule(lr){3-10}
      Pruned (\%) & Full steps & Backg. & Camera & Lang. & Light &
      Layout & Robot & Noise & Avg. & Lat. (ms) & Speedup \\
      \midrule
       0.0 & \(0\)--\(9\) & 66.25 & 51.25 & 87.50 & 95.00 & 83.75 & 51.25 & 61.25 & 70.89 & 428.90 & \(1.00\times\) \\
      \midrule
      68.4& \(0\)       & 57.50 & 31.25 & 85.00 & 92.50 & 81.25 & 46.25 & 47.50 & 63.04 & 207.20 & \(2.07\times\) \\
      53.1 & \(0\)       & 60.00 & 45.00 & 90.00 & 93.75 & 80.00 & 52.50 & 61.25 & 68.93 & 220.47 & \(1.95\times\) \\
      42.9 & \(0\)       & 62.50 & 46.25 & 90.00 & 92.50 & 83.75 & 52.50 & 62.50 & 70.00 & 220.97 & \(1.94\times\) \\
      30.6 & \(0\)       & 63.75 & 48.75 & 90.00 & 88.75 & 85.00 & 60.00 & 68.75 & \textbf{72.14} & 233.02 & \(1.84\times\) \\
      \midrule
      42.9 & \(0,5\)     & 65.00 & 47.50 & 90.00 & 91.25 & 82.50 & 55.00 & 62.50 & 70.54 & 248.80 & \(1.72\times\) \\
      42.9 & \(0,3,6\)   & 61.25 & \textbf{51.25} & 91.25 & 93.75 & 82.50 & 57.50 & 66.25 & 71.96 & 277.04 & \(1.55\times\) \\
      \bottomrule
    \end{tabular}\par
  \end{swamrevision}
\end{table}
\section{Limitations and Future Work}
\label{lab: limitation and future work}

Future work could broaden the evaluation and deployment of
\method{} in three directions. First, our inference benchmarks
were conducted on an NVIDIA RTX 4090; extending evaluation to
other GPUs and resource-constrained devices would help assess
its efficiency across hardware platforms~\citep{wang2026twinquant,wang2025d2moe}. Second, we evaluated
three WAMs, and further studies could examine its applicability
to a broader range of WAM architectures and backbones.
Third, the performance degradation observed under environmental
perturbations motivates adaptive token-selection mechanisms
that adjust to changing conditions to improve robustness while
maintaining inference efficiency.

\section{Conclusion}
\label{sec:conclusion}

In this paper, we presented \method{}, a training-free
method that prunes evolving imagination tokens online
during joint visual-action denoising.
Evaluations on Cosmos 3 Edge, Cosmos 3 Nano Policy,
and FastWAM-Joint across LIBERO, RoboLab-120, and
real-world manipulation tasks demonstrate its
inference efficiency while largely preserving task success.
\method{} achieves approximately $1.8\times$ speedup
on RoboLab-120 and $2.0\times$ on LIBERO.
These results indicate that dense computation over evolving
imagination tokens in WAMs is partially redundant for action
prediction. Sparsifying this computation enables more efficient
inference while largely preserving task performance.

\bibliography{references}
\bibliographystyle{iclr2027_conference}

\appendix

\section{Additional Method Details}
\label{app:method-details}

This section supplements the token selection procedure in
\autoref{subsec:mask-construction} and the attention profiling
mechanism in \autoref{subsec:efficient-execution}.
All attention scores below are obtained at a full denoising
step $\tau_d$.
We omit this step index for brevity and write
$U_\ell(i,f,j)=U_\ell^{(\tau_d)}(i,f,j)$.

\subsection{Token Selection Details}
\label{app:selection-details}

\paragraph{Frame-Aligned Action Queries.}
The temporal alignment observed in
\autoref{fig:observation}, Insight~1(a), motivates aggregating
attention from the action queries associated with each
future frame.
Assuming that $H$ is divisible by $F$, we partition the
$H$ action queries into $F$ consecutive groups:
\begin{equation}
    \mathcal{I}_f
    =
    \left\{
        (f-1)\frac{H}{F}+1,\ldots,f\frac{H}{F}
    \right\},
    \qquad f=1,\ldots,F.
    \label{eq:action-frame-group}
\end{equation}
The frame-aligned score $S_\ell(f,j)$ is obtained by
aggregating $U_\ell(i,f,j)$ over $i\in\mathcal{I}_f$,
as defined in \autoref{eq:frame-aligned-score}.
The query weights satisfy
$\alpha_{f,i}\geq 0$ and
$\sum_{i\in\mathcal{I}_f}\alpha_{f,i}=1$.
Queries near the center of each group receive higher
weights than those near its boundaries, since boundary
queries also attend to neighboring future frames.
The groups themselves remain disjoint.

\paragraph{Layer Quality.}
We assess each layer using both the attention mass assigned
to the associated future frames and the spatial concentration
of that attention.
The frame-aligned attention mass is
\begin{equation}
    R_\ell
    =
    \frac{1}{F}
    \sum_{f=1}^{F}
    \sum_{j=1}^{N_s}
    S_\ell(f,j).
    \label{eq:attention-mass}
\end{equation}
Because $S_\ell(f,j)$ retains the original attention
normalization over all visible keys, $R_\ell$ measures
how much attention is assigned to the associated future
frames.

To measure spatial concentration, we normalize the scores
within each frame:
\begin{equation}
    \bar{S}_\ell(f,j)
    =
    \frac{S_\ell(f,j)}
    {\sum_{j'=1}^{N_s}S_\ell(f,j')}.
    \label{eq:normalized-spatial-score}
\end{equation}
We then compute the normalized spatial
entropy~\citep{zhang2025attention}:
\begin{equation}
    E_\ell
    =
    -\frac{1}{F\log N_s}
    \sum_{f=1}^{F}
    \sum_{j=1}^{N_s}
    \bar{S}_\ell(f,j)\log\bar{S}_\ell(f,j),
    \label{eq:spatial-entropy}
\end{equation}
with the convention $0\log 0=0$.
Lower entropy indicates more concentrated spatial attention.

The resulting layer-quality score is
\begin{equation}
    Q_\ell
    =
    R_\ell(1-E_\ell).
    \label{eq:layer-quality}
\end{equation}
This score favors layers that assign substantial attention
to the associated future frames and localize that attention
within each frame.
We select the $K_{\mathrm{layer}}$ layers with the highest
$Q_\ell$ values to form $\mathcal{L}_{\mathrm{core}}$.
Their scores are aggregated using
\autoref{eq:core-score} to obtain $V(f,j)$ for
frame-specific core selection.

\paragraph{Anchor Candidates and Availability.}
The common candidate pool $\mathcal{P}$ and the core
positions $\mathcal{C}_f$ are defined in
\autoref{eq:core-candidate-pool} and
\autoref{eq:core-selection}, respectively.
Anchors are selected from positions unused by every
frame's core selection:
\begin{equation}
    \mathcal{E}
    =
    \mathcal{J}
    \setminus
    \bigcup_{f=1}^{F}\mathcal{C}_f,
    \qquad
    \mathcal{J}=\{1,\ldots,N_s\}.
    \label{eq:anchor-candidates}
\end{equation}

Since $\mathcal{C}_f\subseteq\mathcal{P}$ for every frame
and $|\mathcal{P}|=N_s-K_s$, we have
\begin{equation}
    |\mathcal{E}|
    =
    N_s-
    \left|
        \bigcup_{f=1}^{F}\mathcal{C}_f
    \right|
    \geq
    N_s-|\mathcal{P}|
    =
    K_s.
    \label{eq:anchor-availability}
\end{equation}
Thus, at least $K_s$ anchor candidates remain available
regardless of how core positions vary across frames.
The candidate set includes all positions outside
$\mathcal{P}$ and any positions inside $\mathcal{P}$
that no frame selects as core.

\paragraph{Shared-Anchor Scores.}
Core localization uses selected layers to emphasize
spatially concentrated attention.
For shared anchors, we aggregate scores from all layers
using the same quality weights:
\begin{equation}
    V_{\mathrm{all}}(f,j)
    =
    \frac{
        \sum_{\ell}Q_\ell S_\ell(f,j)
    }{
        \sum_{\ell}Q_\ell
    }.
    \label{eq:all-layer-score}
\end{equation}
This aggregation includes contextual attention beyond
the layers selected for core localization.

For each spatial position $j$, we compute the mean
and coefficient of variation across future frames:
\begin{equation}
    \begin{aligned}
        \mu_j
        &=
        \frac{1}{F}
        \sum_{f=1}^{F}V_{\mathrm{all}}(f,j),
        \\
        \mathrm{CV}_j
        &=
        \frac{
            \operatorname{Std}_{f}
            \left[V_{\mathrm{all}}(f,j)\right]
        }{\mu_j}.
    \end{aligned}
    \label{eq:anchor-statistics}
\end{equation}
Here, $\operatorname{Std}_{f}$ denotes the standard
deviation across future frames.
The anchor score $\mu_j/(1+\mathrm{CV}_j)$ rewards
strong attention and penalizes variation across frames.
The $K_s$ highest-scoring positions in $\mathcal{E}$
form $\mathcal{S}$, as defined in
\autoref{eq:stable-selection}.

Each frame retains $\mathcal{C}_f\cup\mathcal{S}$.
Because $\mathcal{S}\subseteq\mathcal{E}$,
the core and anchor sets are disjoint in every frame.
Together with $K_c+K_s\leq N_s$, this construction
provides exactly $K_c+K_s$ retained tokens per frame.
The shared set $\mathcal{S}$ specifies spatial positions;
the corresponding token representations remain
frame-specific.

\subsection{Attention Profiling in Pilot}
\label{app:profiling-details}

Pilot computes the scores required for token selection
during the dense conditional forward at each full step.
It evaluates only the relevant action-to-future
query--key pairs and reuses the original attention
normalizers, avoiding an additional network forward
and storage of the full attention matrix.

\paragraph{Preserving the Original Attention Normalization.}
Let $x_{\ell,h}(i,f,j)$ denote the attention logit
between action query $i$ and future-frame key $(f,j)$
at layer $\ell$ and head $h$.
Let $\lambda_{\ell,h}(i)$ denote the log-sum-exp
normalizer for the same query, computed over all
visible keys in the original attention operation.
The corresponding attention probability is therefore
$\exp(x_{\ell,h}(i,f,j)-\lambda_{\ell,h}(i))$.

Pilot combines head averaging and weighted query
aggregation directly:
\begin{equation}
    S_\ell(f,j)
    =
    \frac{1}{N_h}
    \sum_{i\in\mathcal{I}_f}
    \sum_{h=1}^{N_h}
    \alpha_{f,i}
    \exp\!\left(
        x_{\ell,h}(i,f,j)
        -
        \lambda_{\ell,h}(i)
    \right).
    \label{eq:efficient-score-extraction}
\end{equation}
This expression recovers the frame-aligned score
in \autoref{eq:frame-aligned-score}.
Retaining the original normalizers also preserves
the attention mass used to compute $R_\ell$.
Renormalizing only over future-frame keys would
discard their attention mass relative to other
visible tokens and change the layer-quality scores.

\paragraph{Restricted Computation and Direct Reduction.}
For each future frame $f$, the scorer evaluates
query--key products only for action queries in
$\mathcal{I}_f$.
It follows the backbone's query--key normalization,
positional encoding, attention scaling, head mapping,
and visibility constraints when computing these logits.

The computation is batched across future frames,
and the resulting probabilities are reduced directly
over heads and action queries to produce $S_\ell(f,j)$.
Consequently, the scorer stores only the aggregated
spatial scores required for selection, without
materializing the full attention matrix or computing
an additional value-weighted attention output.
Pilot then uses these scores to construct the retained
positions for subsequent sparse steps, as described
in \autoref{subsec:efficient-execution}.

\section{Experimental Details}\label{app:experimental-details}

\subsection{Benchmarks and Evaluation Protocols}
\label{lab: benchmarks}
\paragraph{LIBERO.}
We evaluate on four LIBERO suites~\citep{liu2023libero}: Spatial, Object, Goal, and Long, which cover spatial relationships, object variations, task goals, and long-horizon manipulation, respectively. Each suite contains 10 tasks. We perform 50 rollouts per task, yielding 500 rollouts per suite and 2,000 rollouts across all 40 tasks for each evaluation configuration.

\paragraph{RoboLab-120.}
RoboLab-120~\citep{yang2026robolab} comprises 120 language-conditioned manipulation tasks covering visual, relational, and procedural competencies. The tasks are divided into three difficulty levels: 64 Simple, 39 Moderate, and 17 Complex tasks. We evaluate all 120 tasks with 10 rollouts per task, resulting in 1,200 rollouts per evaluation configuration.

\paragraph{LIBERO-Plus.}
We evaluate robustness using a balanced subset of LIBERO-Plus~\citep{fei25libero-plus} constructed from the same 40 base tasks across the four LIBERO suites. The subset covers seven perturbation categories: background textures, camera viewpoints, language instructions, lighting conditions, object layouts, robot initial states, and sensor noise. For each base task and perturbation category, we select two distinct perturbation variants and perform one rollout per variant. This yields 560 rollouts per evaluation configuration, with 80 rollouts per category and 140 rollouts per suite. All compared methods use the same selected variants, initial-state indices, and environment and model seeds.

\subsection{Implementation Details}
\label{app:audit-implementation}

\paragraph{Backbone architectures.}
Cosmos 3 Nano Policy and Cosmos 3 Edge comprise an understanding tower and a diffusion generation tower. The generation tower jointly denoises future visual and action tokens within a shared Transformer backbone, conditioned on representations from the understanding tower. We cache and reuse instruction representations across action chunks, while updating visual conditioning with each incoming observation. Both models receive one wrist-camera view and two smaller third-person views. FastWAM-Joint uses a Mixture-of-Transformers (MoT) architecture with separately parameterized video and action diffusion Transformers coupled through shared attention. Language embeddings condition both branches through cross-attention, and visual observations comprise one third-person view and one wrist-camera view. Table~\ref{tab:backbone_configs} summarizes the inference configurations.

\begin{table}[H]
\centering
\caption{Backbone configurations used in our simulation experiments.}
\label{tab:backbone_configs}
\small
\vspace{-5pt}
\setlength{\tabcolsep}{4pt}
\renewcommand{\arraystretch}{1.1}
\begin{tabular}{@{}lccc@{}}
\toprule
Configuration & Cosmos 3 & Cosmos 3 & FastWAM- \\
              & Nano Policy & Edge & Joint \\
\midrule
Transformer layers      & 36 & 28 & 30 \\
Camera views            & 3  & 3  & 2  \\
Future frame token      & 360 & 340 & 98 \\
Action horizon          & 32 & 32 & 32 \\
Future latent frames    & 8  & 8  & 2  \\
Denoising steps          & 4  & 4  & 10 \\
Guidance scale          & 3.0 & 3.0 & 1.0 \\
Noise schedule shift    & 5.0 & 5.0 & 5.0 \\
\bottomrule
\end{tabular}
\end{table}

\paragraph{Parameter Setup.}
Table~\ref{tab:method_configs} summarizes the default configurations of \method{} for the three backbones. We use fixed token budgets for each backbone, while selecting informative layers and token positions online for each action chunk. In the evaluated schedule, attention statistics are collected from all Transformer layers during the initial dense conditional forward of each action chunk. Core scores aggregate the six layers with the highest quality scores $Q_\ell$, whereas stable-anchor scores aggregate all layers.

For temporal aggregation, Table~\ref{tab:method_configs} reports the number of aligned queries per frame and the center/outer query weight ratio. Each future latent frame is associated with four action queries in Cosmos 3 and sixteen in FastWAM-Joint. We set the center/outer query weight ratio to 2 for all backbones, assigning twice the weight to queries in the central half of each group compared with those in the outer quarters. After normalization, the weights are $(1,2,2,1)/6$ for Cosmos 3. For FastWAM-Joint, each central query receives weight $1/12$ and each outer query receives weight $1/24$.

\vspace{-8pt}
\begin{table}[H]
\centering
\caption{Default \method{} configurations.}
\label{tab:method_configs}
\small
\vspace{-5pt}
\setlength{\tabcolsep}{4pt}
\renewcommand{\arraystretch}{1.1}
\begin{tabularx}{\columnwidth}{@{}>{\raggedright\arraybackslash}Xccc@{}}
\toprule
Parameter & Nano & Edge & Joint \\
\midrule
Core tokens $K_c$                  & 80  & 80  & \swamrev{28} \\
Stable anchors $K_s$               & 104 & 104 & \swamrev{28} \\
Retained tokens $K_c+K_s$          & 184 & 184 & \swamrev{56} \\
Informative layers $K_{\mathrm{layer}}$ & 6 & 6 & 6 \\
Aligned queries per frame         & 4   & 4   & 16 \\
Center/outer query weight ratio   & 2   & 2   & 2 \\
CV penalty coefficient            & 1.0 & 1.0 & 1.0 \\
$\tau_{d}$ &   
0 & 0 & 0\\
\bottomrule
\end{tabularx}
\end{table}

\subsection{Comparative methods}
\label{app:audit-baselines}

\textbf{ToCa (Zou et al., 2025).}
ToCa selectively reuses token features across denoising steps based on token importance and caching history. We apply caching only to future visual tokens, while observation and action tokens remain fully computed. Dense steps refresh the feature cache, whereas cached steps reuse future-token attention residuals and selectively update their MLP features. We use dense steps $\{0,2\}$ for the four-step Cosmos 3 models and $\{0,4,9\}$ for ten-step FastWAM-Joint, with all remaining steps using cached computation.

\textbf{WorldCache~\citep{feng2026worldcache}.}
WorldCache exploits token-trajectory predictability to reuse or extrapolate denoising predictions. We apply its heterogeneous prediction mechanism to both future visual and action outputs, using fixed dense/cache schedules. For Cosmos 3, the first three steps are dense and the final step uses cached predictions. For FastWAM-Joint, the first three and final steps are dense, while the six intermediate steps use cached predictions. Each action chunk maintains its own prediction history.

\textbf{SpecPrune-VLA~\citep{wang2026specprune}.}
SpecPrune-VLA combines historical and current visual information to guide token selection and progressive layer-wise pruning. We adapt its observation-based selection by broadcasting the selected spatial positions to all future latent frames, while retaining all observation and action tokens. The initial conditional forward constructs a layer-wise pruning plan, which is reused in subsequent denoising forwards. Selected future tokens continue through the Transformer, while pruned tokens retain their exit-layer hidden states, restored before the output heads to preserve the complete latent layout.

\textbf{C$^3$ache~\citep{zhao2026c3ache}.}
C$^3$ache reuses denoising residuals across action chunks at matching denoising steps. We extend this mechanism to the complete joint visual--action Transformer stack, while keeping input embeddings, output heads, and sampling updates active. Fully dense refresh chunks alternate with cache-reuse chunks. In reuse chunks, the first two of four steps use cached residuals for Cosmos 3, and the first five of ten steps use cached residuals for FastWAM-Joint

\begin{swamrevision}
\noindent\textbf{ActionCache~\citep{oi2026actioncache}.}
We maintain a cache of up to 10{,}000 entries and derive cache keys
from hidden states that jointly encode the observed frames and the task
instruction. On a cache hit, we resume joint video--action denoising
from the cached intermediate state and run the final two refinement
steps ($N_{\mathrm{hit}}=2$). On a miss, we run the full denoising schedule.
\par
\end{swamrevision}

\subsection{Timing Details}
\label{app:audit-timing}

We benchmark generation latency per action chunk on Cosmos 3 Edge
using an NVIDIA RTX 4090, batch size 1, BF16 precision, and four
denoising steps. All configurations use identical recorded inputs
and random seeds. After five warm-up runs, we collect 30 measurements
per configuration using CUDA events, with GPU synchronization at
both timing boundaries.
The timed region includes observation encoding, initial profiling,
online scoring and token selection, token packing and restoration,
cache operations, and sampler updates. Model loading, compilation
warm-up, output transfers to the CPU, video decoding, RPC, and
simulation are excluded.

In the main experiments, all accelerated methods use torch.compile and CUDA Graphs. \swamrev{We report median latencies by default, except for C\textsuperscript{3}ache,
which reports latency amortized over complete refresh/reuse cycles.
SpecPrune-VLA uses a minimum retention budget of 184 future tokens per frame. For ActionCache, we estimate mean latency by
weighting the profiled hit and forced-miss latencies using the natural
cache hit rate observed during evaluation.}

For \method{}, LSE-based attention scoring
introduces additional operators, and token selection affects
CUDA Graph capture and execution. The runtime costs of scoring
and selection are included in the measurements.
\swamrev{Table~\ref{tab:swam-rev-latency} separates the gains from sparse inference and execution optimizations through an execution-backend
ablation and a matched-backend comparison. In eager mode, torch.compile and CUDA Graphs are disabled. Parenthesized speedups compare Sparse-WAM eager with dense eager and Sparse-WAM with full Pilot with optimized dense.}

\begin{table}[h]
\begin{swamrevision}
  \centering
  \caption{Latency and speedup of \method{} and Pilot across backbones.}
  \label{tab:swam-rev-latency}
  \small
  \begin{tabular}{@{}lrrr@{}}
    \toprule
    Execution configuration & FastWAM-Joint & Cosmos3-Nano-Policy & Cosmos3-Edge \\
    \midrule
    Dense eager (ms) & 434 & 3450 & 859 \\
    \method{} eager (ms)
      & 287 (\(1.51\times\)) & 2144 (\(1.61\times\)) & 556 (\(1.55\times\)) \\
    Optimized Dense (ms) & 346 & 3026 & 727 \\
    \method{} (full Pilot) (ms)
      & 219 (\(1.58\times\)) & 1906 (\(1.59\times\)) & 465 (\(1.56\times\)) \\
    \midrule
    \bottomrule
  \end{tabular}\par
\end{swamrevision}
\end{table}

\subsection{Real World Experiment}
\label{app:fine-tune}

\paragraph{Finetuning.}
We perform full-parameter finetuning on three real-world tasks:
placing three objects from the table into a container,
assembling the mouse battery, and stacking three cups into a small
tower. Training uses eight NVIDIA H800 GPUs with a global batch
size of 64. Each prediction produces an action chunk of 32 steps
conditioned on a single observation, without observation history.
Table~\ref{tab:real_world_finetuning} summarizes the training
configuration and task-specific training budgets.

\begin{table}[H]
\centering
\caption{Finetuning configurations for real-world experiments.}
\label{tab:real_world_finetuning}
\small
\vspace{-5pt}
\begin{tabular}{@{}p{0.62\columnwidth}p{0.30\columnwidth}@{}}
\toprule
Parameter & Value \\
\midrule
Finetuning & Full parameters \\
GPUs & $8\times$ H800 \\
Batch size per GPU & 8 \\
Global batch size & 64 \\
Gradient accumulation & 1 \\
Learning rate & $10^{-4}$ \\
Learning-rate scheduler & Cosine \\
Weight decay & 0.01 \\
Mixed precision & BF16 \\
Gradient clipping & 1.0 \\
Action chunk size & 32 \\
Observation history & Single-step \\
\midrule
Task & Training steps \\
\midrule
Place three objects into a container & 15{,}000 \\
Assemble the mouse battery & 20{,}000 \\
Stack three cups into a tower & 15{,}000 \\
\bottomrule
\end{tabular}
\end{table}

\begin{figure}[H]
    \centering
    \includegraphics[width=\linewidth]{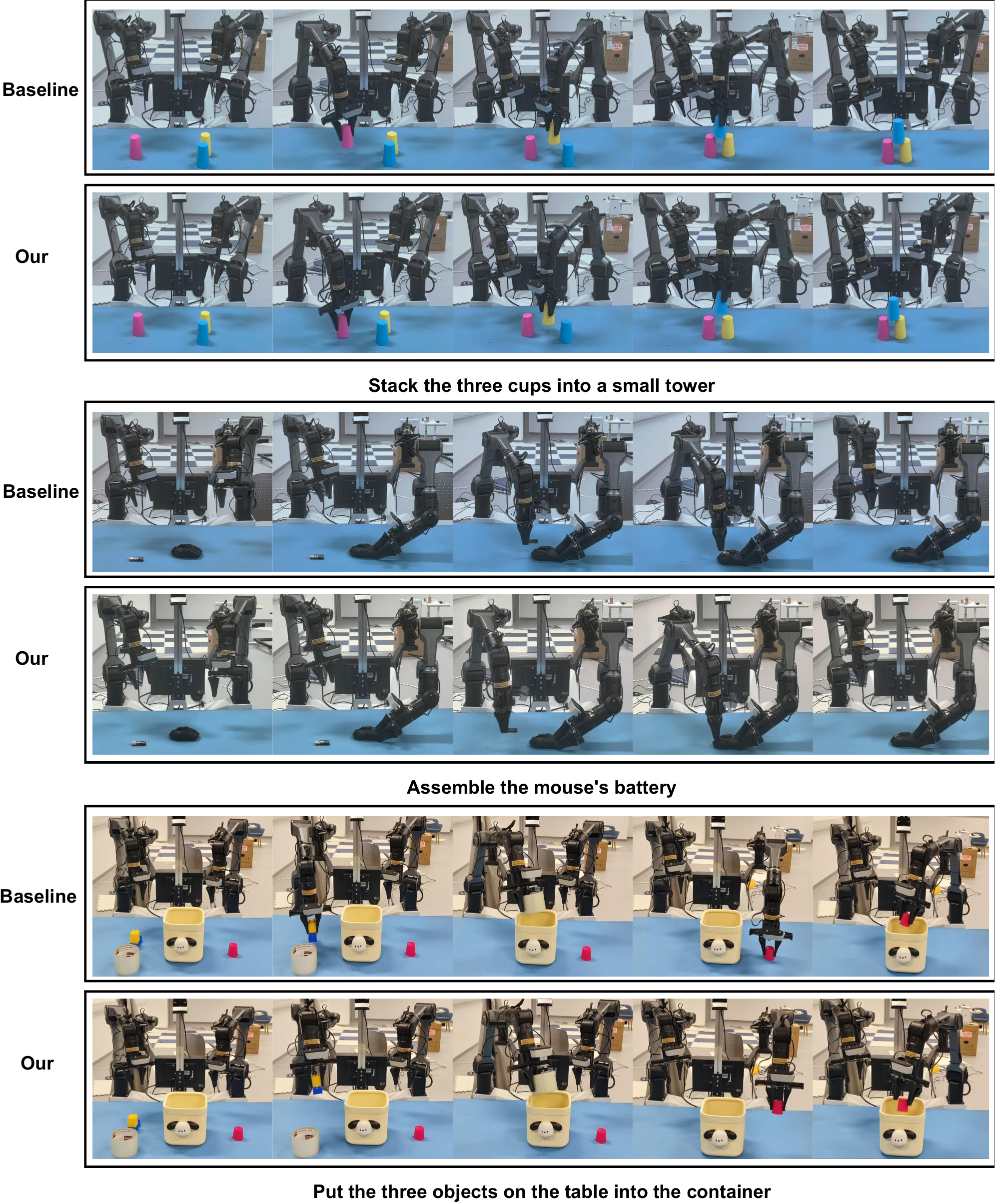}
    \caption{Comparison between baseline and \method{} in real world.}
    \label{fig:real_world_exp}
\end{figure}

\paragraph{Evaluation.}
Real-world evaluation results are presented in
Figure~\ref{fig:real_world_exp}.
We evaluate dense FastWAM-Joint and Sparse-WAM on object packing, cup stacking, and the battery task. Table~\ref{tab:real-world-results} reports the task-level success rates, average success rate, and inference latency.

\section{Extended Experiments} \label{app:extended-experiments}

\subsection{Robustness Analysis}
\label{app:audit-robustness}

\begin{figure}[!htbp]
    \centering
    \includegraphics[width=0.95\linewidth]{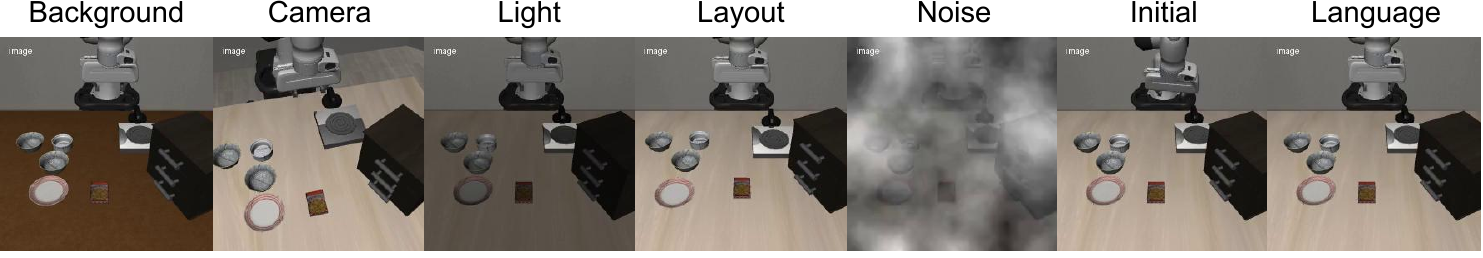}
    \caption{LIBERO-Plus visualization.}
    \label{fig:audit-perturbations}
\end{figure}

\begin{swamrevision}
Table~\ref{tab:swam-rev-robustness} reports the full robustness
sweep. The optimized dense timing reference for this sweep is \(345.23\) ms
per action chunk.
\par
\end{swamrevision}

\subsection{Additional ablations}
\label{app:audit-additional-ablations}

\paragraph{Informative layer aggregation.}
We compare informative-layer and all-layer attention
aggregation for core token selection on Cosmos3-Nano-Policy.
Table~\ref{tab:selection-ablations}(a) shows that aggregating attention from informative layers improves
the average success rate by 2.25 percentage points,
with little change in inference latency.
This result supports selecting informative layers
to guide core token selection.

\paragraph{Temporal action weighting.}

Motivated by the temporal action--frame alignment
observed in Figure~\ref{fig:observation}, Insight~1(a),
we compare uniform and center-weighted action aggregation
for core token scoring.
Center-weighted aggregation emphasizes queries near
the center of each frame-associated action group,
which attend more strongly to the corresponding frame.
As shown in Table~\ref{tab:selection-ablations}(b),
this weighting improves the average success rate
from 32.92\% to 35.83\%, supporting the use of temporal
alignment when aggregating action-to-future attention.
\begin{table}[!htbp]
    \centering
    \caption{
    Ablations of informative-layer aggregation and temporal action weighting on Cosmos 3 Nano Policy. Panel (a) compares all-layer and informative-layer aggregation; panel (b) compares uniform and center-weighted action aggregation.
    }
    \label{tab:selection-ablations}
    \small
    \vspace{-5pt}
    \setlength{\tabcolsep}{4pt}
    \renewcommand{\arraystretch}{1.1}

    \begin{minipage}[t]{0.49\linewidth}
    \vspace{0pt}
    \centering
    \textbf{(a) Informative layer selection}
    \par\smallskip
    \small
    \setlength{\tabcolsep}{3pt}

    \begin{tabularx}{\linewidth}{@{}Xrrr@{}}
            \toprule
            Core aggregation
            & \shortstack{Avg. SR(\%)}
            & Score 
            & \shortstack{latency(ms) } \\
            \midrule
            All layers
            & 33.25 & 48.80 & 1846.8 \\
            Informative layers
            & \textbf{35.50} & \textbf{51.84} & 1857.4 \\
            \bottomrule
        \end{tabularx}
        \label{tab:block-aggregation}
    \end{minipage}
    \hfill
    \begin{minipage}[t]{0.49\linewidth}
        \vspace{0pt}
        \centering
        \textbf{(b) Temporal action weighting}
        \par\smallskip

        \begin{tabularx}{\linewidth}{@{}Xrr@{}}
            \toprule
            Action weighting
            & \shortstack{Avg. SR(\%)}
            & Score  \\
            \midrule
            Uniform
            & 32.92 & 48.31 \\
            Center-weighted
            & \textbf{35.83} & \textbf{50.71} \\
            \bottomrule
        \end{tabularx}
        \label{tab:temporal-action-weighting}
    \end{minipage}
\end{table}

\begin{swamrevision}
\paragraph{Component ablation on FastWAM-Joint.}
On LIBERO, keeping the same sparsity ratio across all sparse variants,
combining core tokens and stable anchors achieved the highest mean
success rate among them (\(98.45\%\), compared with \(97.60\%\) for
core-only and \(98.10\%\) for stable-only selection) without incurring
substantially higher computational overhead, as all four sparse variants
achieved similar speedups (Table~\ref{tab:fastwam-components}). All sparse variants prune 42.9\% of future tokens and use the same full Pilot configuration. Speedups are relative to dense eager inference.
\par
\end{swamrevision}

\begin{table}[!htbp]
  \begin{swamrevision}
    \centering
    \caption{Token selection ablation on LIBERO with FastWAM-Joint.}
    \label{tab:fastwam-components}
    \small
    \begin{tabular}{@{}lrrrrrr@{}}
      \toprule
      & \multicolumn{5}{c}{Success rate (\%)} & \\
      \cmidrule(lr){2-6}
      Selection & Spatial & Object & Goal & Long & Avg. & Speedup \\
      \midrule
      Dense eager         & 99.00 & 100.00 & 98.20 & 97.80 & 98.75 & \(1.000\times\) \\
      \midrule
      Core-only     & 98.00 &  99.60 & 96.80 & 96.00 & 97.60 & \(1.977\times\) \\
      Stable-only   & 99.60 &  99.20 & 96.80 & 96.80 & 98.10 & \(1.983\times\) \\
      Random        & 98.80 &  98.80 & 95.20 & 92.40 & 96.30 & \(1.978\times\) \\
      Core + Stable & 98.80 &  99.20 & 98.60 & 97.20 & \textbf{98.45} & \(1.985\times\) \\
      \bottomrule
    \end{tabular}\par
  \end{swamrevision}
\end{table}

\begin{swamrevision}
\paragraph{Pruning ratio on FastWAM-Joint.}
We split the retained budget equally between core tokens and stable
anchors. Reducing the per-frame future-token budget yields a clear
speed--accuracy trade-off: it increases speedup over optimized dense
inference but lowers the mean success rate (Table~\ref{tab:fastwam-pruning}). Latencies and speedups use timing measurements from this pruning-ratio sweep.
\par
\end{swamrevision}

\begin{table}[!htbp]
  \begin{swamrevision}
    \centering
    \caption{Pruning-ratio ablation on LIBERO with FastWAM-Joint.}
    \label{tab:fastwam-pruning}
    \footnotesize
    \setlength{\tabcolsep}{2.5pt}
    \begin{tabular}{@{}rc*{8}{r}@{}}
      \toprule
      & & \multicolumn{5}{c}{Success rate (\%)}
      & & \multicolumn{2}{c}{Speedup vs.\ dense} \\
      \cmidrule(lr){3-7}\cmidrule(lr){9-10}
      Pruned (\%) & Core/Stable
      & Spatial & Object & Goal & Long & Avg. & Lat. (ms)
      & Eager & Optimized \\
      \midrule
       0.0 & --    & 99.00 & 100.00 & 98.20 & 97.80 & 98.75 & --
         & \(1.00\times\) & \(1.00\times\) \\
      \midrule
      71.4 & 14/14 & 96.40 &  98.00 & 98.40 & 94.60 & 96.85 & 207
         & \(2.10\times\) & \(1.67\times\) \\
      51.0 & 24/24 & 98.00 & 100.00 & 97.60 & 96.40 & 98.00 & 220
         & \(1.96\times\) & \(1.57\times\) \\
      42.9 & 28/28 & 98.80 &  99.20 & 98.60 & 97.20 & 98.45 & 219
         & \(1.97\times\) & \(1.58\times\) \\
      30.6 & 34/34 & 98.80 & 100.00 & 97.00 & 97.80 & 98.40 & 234
         & \(1.85\times\) & \(1.48\times\) \\
      \bottomrule
    \end{tabular}\par
  \end{swamrevision}
\end{table}

\subsection{Selection Reuse and Refresh Frequency}
\label{app:audit-selection-reuse}

\begin{figure}[!htbp]
    \centering
    \includegraphics[width=0.95\linewidth]{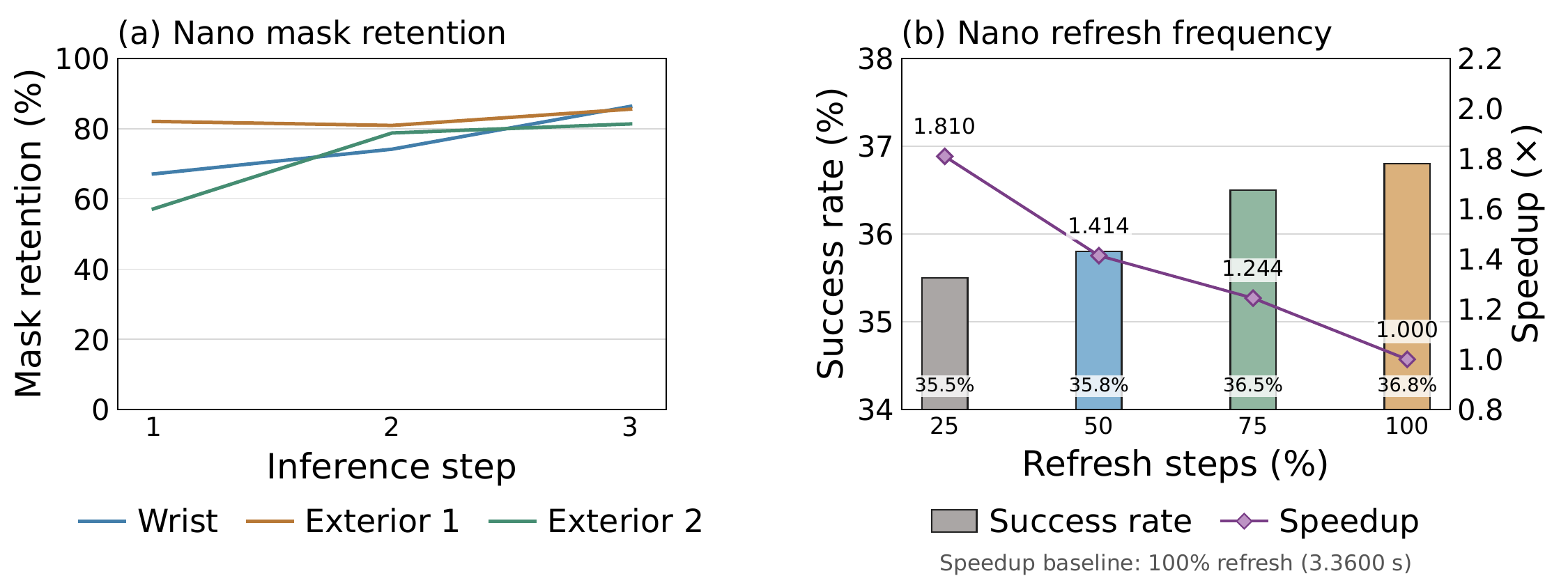}
    \caption{Selection reuse on Cosmos 3 Nano Policy.
    (a) Mask retention across denoising steps for three camera views.
    (b) Success rate and speedup under different proportions of
    full-refresh steps. Speedup is measured relative to full refresh
    at every denoising step.}
    \label{fig:selection-reuse}
\end{figure}

We examine the effectiveness of reusing token selections
across denoising steps on Cosmos 3 Nano Policy.
Figure~\ref{fig:selection-reuse}(a) reports mask retention
across three camera views, showing substantial retention
in later denoising steps.
This observation motivates reusing selected spatial
positions rather than repeatedly recomputing the selection.

We further vary the proportion of full-refresh steps,
which perform dense computation and update token selections,
while the remaining steps reuse the selections for
sparse inference.
As shown in Figure~\ref{fig:selection-reuse}(b),
a single full-refresh step in the four-step denoising
schedule achieves a $1.81\times$ speedup with a
35.5\% success rate, compared with 36.8\% under
full refresh at every step.
More frequent refreshes improve success rate
but reduce the acceleration benefit.
These results support using limited dense scoring
and cross-step selection reuse to achieve substantial
acceleration with a modest reduction in task success.

\subsection{CROSS-STEP AND CROSS-FRAME ATTENTION OVERLAP}
\label{app:audit-overlap}

We examine action-to-future attention on Cosmos 3 Edge
and FastWAM-Joint to assess its consistency across
denoising steps and future frames.
For two normalized spatial attention distributions
$P$ and $Q$, we measure their overlap as
\begin{equation}
    \operatorname{Overlap}(P,Q)
    = \sum_i \min(P_i,Q_i),
    \label{eq:attention-overlap}
\end{equation}
where higher values indicate greater agreement
in the spatial distribution of attention.

\begin{figure}[!htbp]
    \centering
    \includegraphics[width=\linewidth]{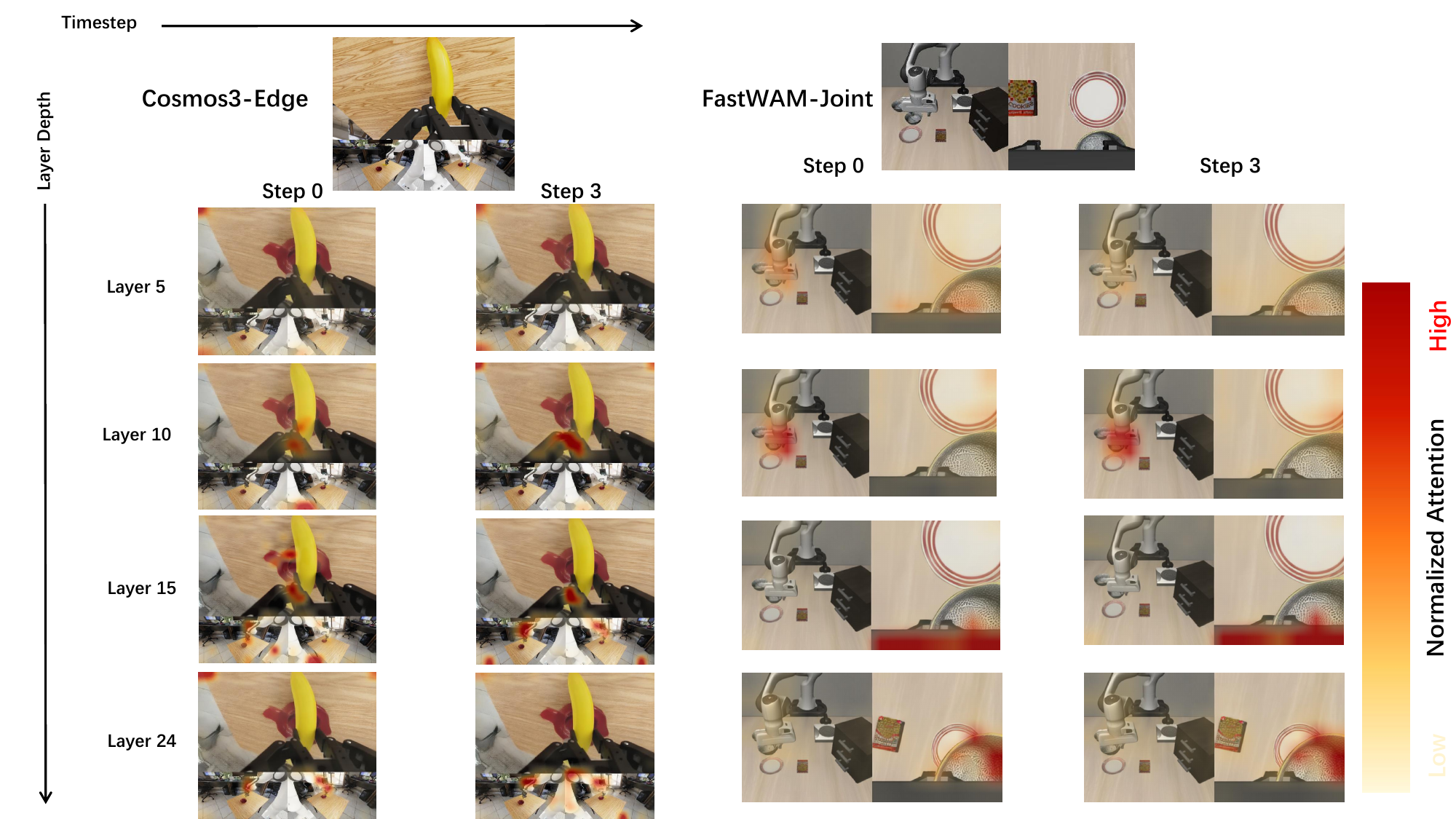}
    \caption{Visualization of action-to-future attention on Cosmos3 Edge and Fastwam-Joint}
    \label{fig:edge_fastwam}
\end{figure}

\paragraph{Cross-step consistency.}
 Using the overlap measure in Equation~\ref{eq:attention-overlap}, we measured mean overlap between consecutive denoising steps of 81.11\% on Cosmos 3 Edge and 97.97\% on FastWAM-Joint. Figure~\ref{fig:edge_fastwam} provides representative visualizations of action-to-future attention. The measured agreement in spatial attention distributions motivates selection reuse between full steps.

\paragraph{Cross-frame variation.}
Mean cross-frame overlap is lower, reaching 68.89\%
on Cosmos 3 Edge and 81.85\% on FastWAM-Joint.
Within each model, this contrast indicates greater
variation in attention across future frames than
across adjacent denoising steps.
Such variation motivates selecting core tokens
independently for each future frame, allowing
spatial selections to follow changes in
action-relevant regions along the predicted future.

\newpage
\subsection{Experiment Visualization}
\label{app:audit-visualization}
\begin{figure}[t]
    \centering
    \includegraphics[width=\linewidth]{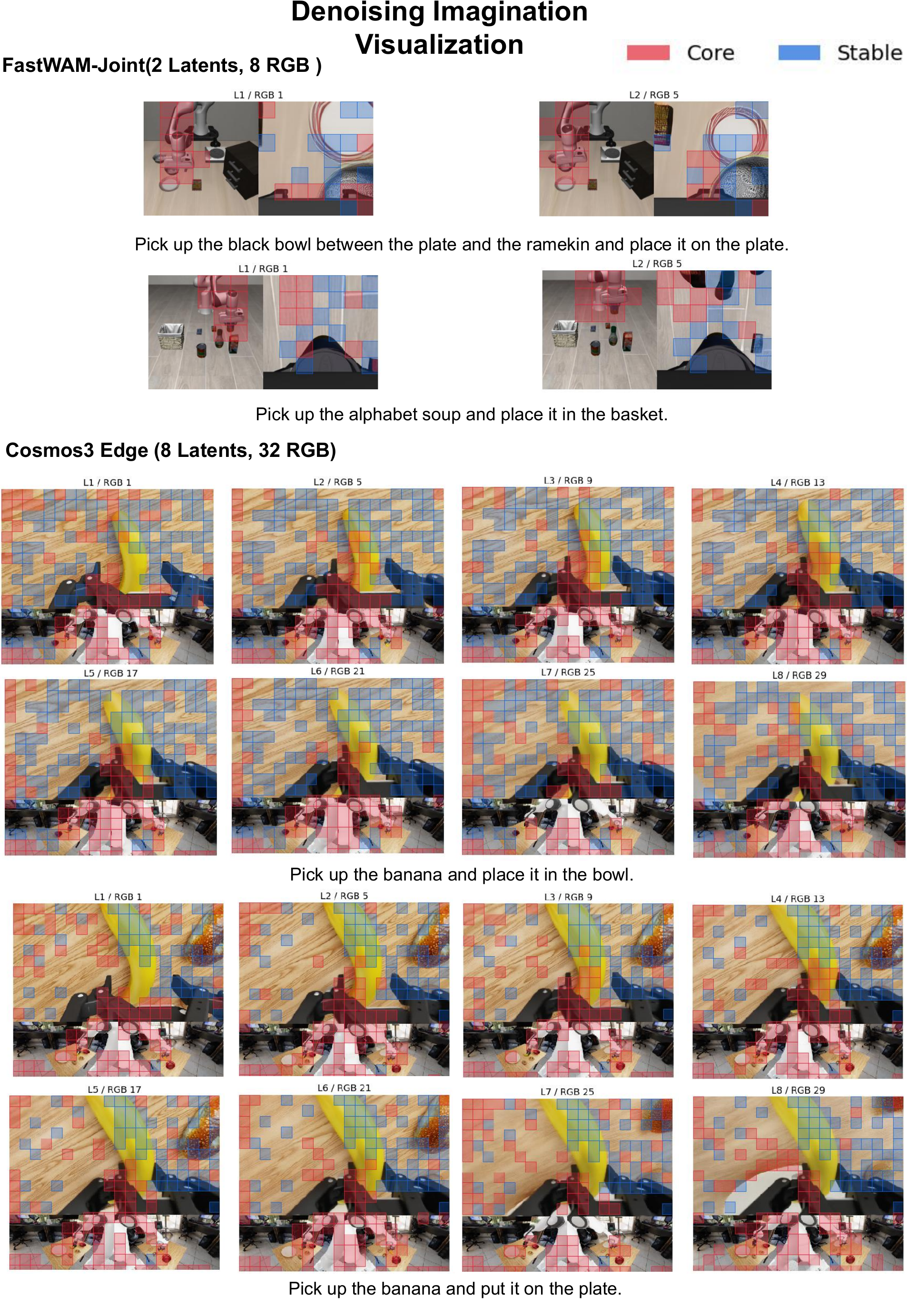}
    \caption{Core-token and stable-anchor selections in imagined
    future frames from FastWAM-Joint and Cosmos 3 Edge.
    Red and blue patches denote frame-specific core tokens and
    shared spatial anchors, respectively.}
    \label{fig:imagination-visualization}
\end{figure}

To examine which regions of imagined futures are
retained by \method{}, Figure~\ref{fig:imagination-visualization}
visualizes token selections on four manipulation tasks
using FastWAM-Joint and Cosmos 3 Edge.
Red patches denote frame-specific core tokens,
while blue patches denote stable anchors at shared
spatial positions across future frames.
The visualizations illustrate how \method{} retains
object--gripper interaction regions together with
surrounding spatial context, combining frame-specific
attention hotspots with shared contextual anchors.

\par\vfill\newpage

\end{document}